\documentclass{article}
\usepackage{ijcai23}

\usepackage{times}
\usepackage{soul}
\usepackage{url}
\usepackage[hidelinks]{hyperref}
\usepackage[utf8]{inputenc}
\usepackage[small]{caption}
\usepackage{graphicx}
\usepackage{amsmath}
\usepackage{amsthm}
\usepackage{booktabs}
\usepackage{algorithm}
\usepackage{algorithmic}
\usepackage{subfigure}
\usepackage{makecell}

\newtheorem{theorem}{Theorem}

\newtheorem{definition}{Definition}

\usepackage{amsfonts}       
\usepackage{nicefrac}       
\usepackage{microtype}      

\usepackage{times}
\usepackage{soul}
\usepackage{fdsymbol}
\usepackage{graphicx}
\usepackage{amsmath}

\usepackage{import}
\usepackage{subfigure}

\usepackage{xcolor}

\labelformat{equation}{(#1)}

\title{Structural Hawkes Processes for Learning Causal Structure from Discrete-Time Event Sequences}

\author{
Jie Qiao$^1$\and
Ruichu Cai$^{1,2}$\thanks{Corresponding author.}\and
Siyu Wu$^{1}$\and
Yu Xiang$^1$\and
Keli Zhang$^3$\And
Zhifeng Hao$^4$
\affiliations
$^1$School of Computer Science, Guangdong University of Technology, Guangzhou 510006, China\\
$^2$Peng Cheng Laboratory, Shenzhen 518066, China\\
$^3$Huawei Noah’s Ark Lab, Huawei, Shenzhen 518116, China\\
$^4$College of Science, Shantou University, Shantou 515063, China
\emails
qiaojie.chn@gmail.com,cairuichu@gmail.com,fisherwsy@163.com,
thexiang2000@gmail.com,zhangkeli1@huawei.com,haozhifeng@stu.edu.cn
}

\begin{document}

\maketitle

\begin{abstract}
    Learning causal structure among event types from discrete-time event sequences is a particularly important but challenging task. Existing methods, such as the multivariate Hawkes processes based methods, mostly boil down to learning the so-called Granger causality which assumes that the cause event happens strictly prior to its effect event. Such an assumption is often untenable beyond applications, especially when dealing with discrete-time event sequences in low-resolution; and typical discrete Hawkes processes mainly suffer from identifiability issues raised by the instantaneous effect, i.e., the causal relationship that occurred simultaneously due to the low-resolution data will not be captured by Granger causality. In this work, we propose Structure Hawkes Processes (SHPs) that leverage the instantaneous effect for learning the causal structure among events type in discrete-time event sequence. The proposed method is featured with the minorization-maximization of the likelihood function and a sparse optimization scheme. Theoretical results show that the instantaneous effect is a blessing rather than a curse, and the causal structure is identifiable under the existence of the instantaneous effect. Experiments on synthetic and real-world data verify the effectiveness of the proposed method.
\end{abstract}

\section{Introduction}

\begin{figure}[t]
    \centering
    \subfigure[Continuous-time event sequences.]{\label{fig:toy_example_a}
    \includegraphics[width=0.45\textwidth]{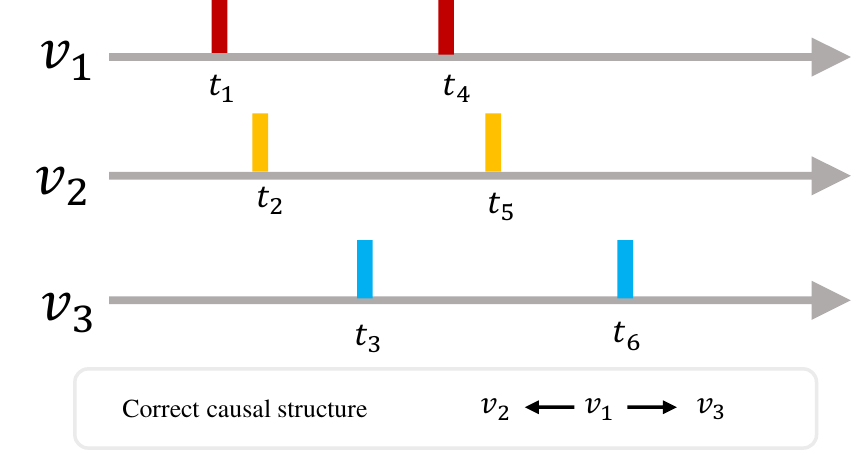}
    }
    \subfigure[Discrete-time event sequences.]{
    \label{fig:toy_example_b}
    \includegraphics[width=0.45\textwidth]{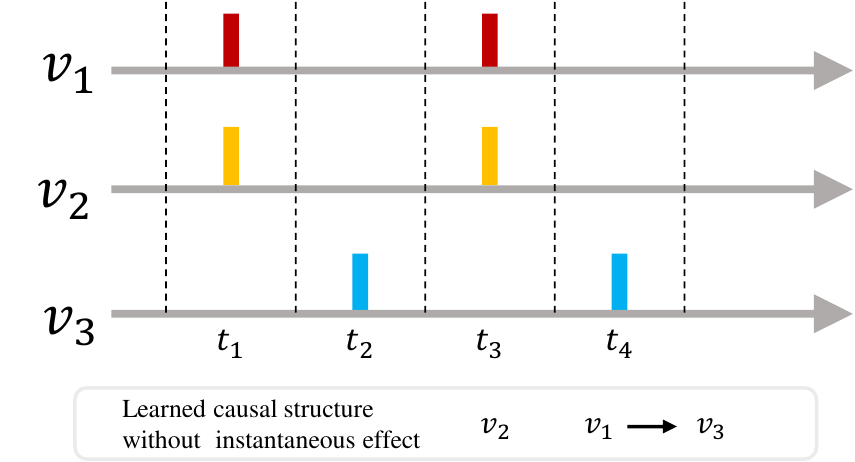}
    }
    \caption{Toy example of the continuous-time and discrete-time event sequences. We consider three types of event $v_1,v_2,v_3$ and the causal relationship satisfying $v_2\leftarrow v_1 \rightarrow v_3$. While, in continuous-time, each event's occur times $(t_1,...,t_6)$ can be observed accurately, the discrete-time event sequences have lower resolution, such that types $v_1,v_2$ will be considered as simultaneous and the Granger-based methods will fail to capture the causal relationship $v_1\to v_2$.}
    \label{fig:toy_example}
\end{figure}

Learning causal structure among event types on \textit{multi-type event sequences} is an important and challenging task, and has recently found applications in social science \cite{zhou2013learningsocial}, economic \cite{bacry2015hawkes}, network operation maintenance \cite{cai2022thps}, etc. Existing methods, such as the multivariate Hawkes processes based methods \cite{xu2016learning,bhattacharjya2018proximal,salehi2019learning}, mostly boil down to learning the so-called Granger causality \cite{granger1969investigating} which implicitly assumes that all events are recorded instantaneously and accurately such that the cause event happens strictly prior to its effect event (known as \textit{temporal precedence assumption}).
However, due to the limited recording capabilities and storage capacities, retaining event's occurred times with high-resolution is expensive or practically impossible in many real-world applications, and we usually only can access the corresponding discrete-time event sequences. For example, in large wireless networks, the event sequences are usually logged at a certain frequency by different devices whose time might not be accurately synchronized.
As a result, low-resolution discrete-time event sequences are obtained and the temporal precedence assumption will be frequently violated in discrete-time event sequences, which raises a serious identifiability issue of causal discovery.
For example, as shown in Fig. \ref{fig:toy_example}, there are three event sequences produced by three event types $v_1$, $v_2$, and $v_3$, respectively. Let $v_1$ be the cause of $v_2$ and $v_3$, Fig. \ref{fig:toy_example_a} shows the accurate continuous-time event sequences such that each event occurred time is recorded by $t_1,...,t_6$. However, such high-resolution sequences are usually not accessible and we can only observe the discrete-time event sequences as shown in Fig. \ref{fig:toy_example_b}. As a result, $v_1$ and $v_2$ will be considered simultaneous events due to the low-resolution, which violates the temporal precedence assumption. Consequently, many existing point process based methods will fail to capture the causal relationship $v_1\to v_2$ as only the events that occur earlier are considered as causes.
In contrast, the causal relationship $v_1\to v_3$ can still be captured using the point process based method because the cause event $v_1$ occurs before $v_3$. However, one can imagine that as the resolution becomes lower the causal relationship between $v_1$ and $v_3$ might no longer be identified as they might become occurs at the same time. Thus, in this paper, we aim to answer the following two questions: 1) How to design and learn a Hawkes process that leverages the instantaneous effect in discrete time? 2) Can we identify the causal relationship in event sequences under the existence of the instantaneous effect?

These two questions have to do with the point processes and the causal discovery, respectively. The former question is about the design of the discrete-time Hawkes processes. The effect of discretization of Hawkes processes has been widely discussed in many ways \cite{foufoula1986continuous,kirchner2016hawkes,shlomovich2022parameter,jacod2009testing}. However, how to design a Hawkes process that leverages the instantaneous effect in discrete Hawkes process is still unknown. The latter question is related to the causal discovery \cite{zhang2018learning,qiao2021causal,cai2018causal,cai2018self,yu2019multi,yu2021unified,yang2021learning}. Some methods have been developed for continuous-value time series with instantaneous effect. For example, the structural vector autoregression model \cite{swanson1997impulse} is an extension of the vector autoregression model with instantaneous effect,
and it has been shown that under the linear non-Gaussian model, the causal relationship is identifiable under the existence of the instantaneous effect \cite{hyvarinen2010estimation}. However, we are not aware of any method that leverages the instantaneous effect to identify the causal relationship in event sequences with point processes. 
Thus, in this work, we propose Structural Hawkes Processes (SHPs) that leverage the instantaneous effect in discrete-time event sequences.
We will proceed as follows. In Section \ref{sec:related work}, we review the related work. In Section \ref{sec:mpp}, we show how to design and learn the structural Hawkes processes. In Section \ref{sec:identifiability}, we investigate the identification of structural Hawkes processes and theoretically show that the causal relationship among the instantaneous effect is indeed identifiable. In Section \ref{sec:experiments}, we perform extensive empirical evaluations over both synthetic and real-world data.

\section{Related Work}\label{sec:related work}

This work is closely related to two topics: point processes and learning causal structure from event sequence. 
\paragraph{Point processes.} 
There are mainly two types of point processes for modeling the event sequence. Most existing methods focus on developing the continues-time Hawkes processes \cite{hawkes1971spectra} which assume that the event in the past can influence the event in the future. Different variants have been developed with different types of intensity in the Hawkes processes, e.g., the parametric functions \cite{farajtabar2014shaping,zhou2013learningsocial,rasmussen2013bayesian,cai2022thps}, the non-parametric functions \cite{lewis2011nonparametric,zhou2013learning,achab2017uncovering}, and the recent deep learning based functions \cite{du2016recurrent,mei2017neural,shang2019geometric}. 
Another line of research focuses on the discrete-time Hawkes processes which are more appealing for certain applications. 
\cite{seol2015limit} study the limit theorem for the discrete Hawkes processes and it has been extended to the discrete marked Hawkes processes by \cite{wang2022limit}. \cite{shlomovich2022parameter} further discusses the estimation method for 1-dimensional discrete Hawkes processes. However, none of the above methods leverage the instantaneous effect for learning causal structure in discrete-time event sequences.

\paragraph{Learning causal structure from event sequences.}
Most approaches in learning causal structure from event sequences are based on Granger causality \cite{granger1969investigating}. The basic idea of Granger causality is to constrain that the effect cannot precede the cause and the causal analysis can be conducted using predictability. In particular, many methods have been developed for learning Granger causality from the continuous-time event sequences based on multivariate Hawkes processes. \cite{xu2016learning} proposes a nonparametric Hawkes processes model with group sparsity regularization, while \cite{zhou2013learningsocial} proposes to use a nuclear and $\ell_1$ norm and \cite{ide2021cardinality} consider an $\ell_0$ norm as the sparse regularization. In addition, \cite{achab2017uncovering} proposes to use cumulant for learning the causal structure without optimization. Recently, some deep point process based methods have been proposed, e.g., \cite{02817a8a3ee244c68c2a91510915c943} introduce an attribution method to uncover the Granger causality. 
However, the Granger causal relations could be misled in low-resolution which is also pointed out by \cite{spirtes2016causal}. One remedy is to extend Granger causality with instantaneous effects. It has been found that in continuous-value time series, one is able to incorporate the instantaneous effect under the linear relation with non-Gaussian noise \cite{hyvarinen2010estimation}. \cite{runge2020discovering} further proposes to use a constraint-based method for learning the causal structure under the instantaneous effects. However, the extensions are only applicable in a restricted case and it is still very challenging to learn causal structure from event sequences with instantaneous effects.

\section{Structural Hawkes Processes}
We begin with a brief introduction to the general continuous-time multivariate point processes. We then develop the structural Hawkes processes that leverage the instantaneous effect in discrete-time. 

\subsection{Multivariate Point Processes}\label{sec:mpp}
A multivariate point process is a random process that can be presented
via a $|\mathbf{V}|$-dimensional counting process $\mathbf{N}=\{N_v(t)|t\in [0,T],v\in \mathbf{V} \}$ where $N_v(t)=N((0,t])$ measures the number of events that occur before time $t$ in the event type $v$. Each counting process $N_v(t)$ can be characterized by the \textit{conditional intensity function} $\lambda_{v}(t)$ satisfying:
\begin{equation*}\label{eq: lambda}
  \lambda_{v}(t)dt =\mathbb{E}[dN_{v}(t)|{\mathcal{F}_{t-}}],  
\end{equation*}
where $\lambda_{v}(t)$ characterizes the (conditional) expected number of jumps per unit of time, $dN_v(t)=N_v(t+dt)-N_v(t)$ measures the increment of the jump, $\mathcal{F}_{t-}=\bigcup_{0\leq s < t,v\in \mathbf{V}}\mathcal{F}^{v}_{s}$ in which $\mathcal{F}^{v}_{s}$ is the canonical filtration of sub-processes $N_v(t)$ up to time $s$.

In particular, Hawkes process is a counting process with an intensity function that has the following form:
\begin{align}
\label{eq:intensity_continue}
\lambda _{v} (t) & =\mu_{v} +\sum _{v'\in \mathbf{V}}\int_{0}^t \phi _{v',v} (t-t')dN_{v'}( t'),
\end{align}
where $\mu_{v}$ is the immigration intensity and $\sum_{v'\in \mathbf{V}}\int _{0}^t \phi _{v',v} (t-t')dN_{v'}( t')$ is the endogenous intensity aiming to model the influence from other event types occurring near time $t$ \cite{farajtabar2014shaping}.
$\phi_{v',v}(t)$ is a reproduction function characterizing the time variation of the causal influence from event $v'$ to $v$. Intuitively, one can imagine that events can be seen to arrive either via immigration according to $\mu_{v}$ or via reproduction from past events according to the endogenous intensity.

\subsection{Design of Structural Hawkes Processes}\label{sec:dthp}

To model the instantaneous effect, we begin with extending the continues-time counting processes into the discrete-time in which the event sequences are observed or collected in $\mathbf{T}=\{\Delta,2\Delta,...,K\Delta\}$, for $K=\lfloor T/\Delta \rfloor$ where $\Delta>0$ is the length of time interval at each observed time. Then the multivariate counting processes in discrete-time can defined as $\mathbf{N}^{(\Delta)}=\{N_v^{(\Delta)}(k)|k\in\{0,\dots,K\},v\in \mathbf{V}\}$, where $N_v^{(\Delta)}(k)=N_v((0,k\Delta])$ measures the number of events that occurs not later than $k\Delta$. We further let $\mathbf{X}=\{X_{v,t}|v\in \mathbf{V},t\in \{0,\dots,K\}]\}$ denote the set of observations at each time interval where $X_{v,t}\coloneq N_{v}( t\Delta)-N_{v}( (t-1)\Delta)$ is an analogy to $dN_v(t)$.

The discrete-time counting processes, however, as we discussed earlier, will ignore the instantaneous effect which could lead to a misleading result. To tackle this issue, we propose structural counting processes with a new type of conditional intensity function that leverages the events that occur at the same period of time.
\begin{definition}[Structural counting processes]\label{def:scp}
A structural counting process is a multivariate counting process $\mathbf{N}^{(\Delta)}$ in discrete-time with the conditional intensity of $N^{(\Delta)}_v$ for each $v\in \mathbf{V}$ satisfying:
\begin{equation}\label{eq:lambda_instant}
  \lambda_{v}(k\Delta)\Delta =\mathbb{E}[X_{v,k}|{\mathcal{F}_{(k-1)\Delta}\cup \mathcal{F}^{-v}_{k\Delta}}],  
\end{equation}
where $\mathcal{F}_{(k-1)\Delta } =\bigcup_{0 \leq s \leq k-1,v\in \mathbf{V}}\mathcal{F}^{v}_{s\Delta }$ is the filtration with discrete-time in the past and $\mathcal{F}^{-v}_{k\Delta } \coloneq \left\{\mathcal{F}^{v'}_{k\Delta } |v'\in \mathbf{V} \backslash v\right\}$ is the filtration that except for type-$v$ event.
\end{definition}
Note that the exclusion in $\mathcal{F}^{-v}_{k\Delta }$ is necessary since it makes no sense to use the current number of events to predict itself. Based on the structural counting processes in Definition \ref{def:scp}, the structural Hawkes processes can be designed as follows:
\begin{definition}[Structural Hawkes processes]
A structural Hawkes process is a structural counting process such that for each $v\in \mathbf{V}$, the intensity of $N^{(\Delta)}_v$ can be written as:
\begin{equation}\label{eq:discrete_intensity_continue_instant}
\lambda _{v} (k\Delta) =\mu_{v} +\sum _{v'\in \mathbf{V}}\sum_{i=1}^{k} \phi _{v',v} ((k-i)\Delta)X_{v',i},
\end{equation}
where $\phi_{v,v}(0)\equiv 0$ ensures the exclusive of type-$v$ event at time $k\Delta$.
\end{definition}
We can see that the intensity in Eq. \ref{eq:discrete_intensity_continue_instant} is not only influenced by the events that occur in the past $(k-1)\Delta$ but also the events that occur at the same period of time $k\Delta$. 

\subsection{Learning of Structural Hawkes Processes}

The goal of this work is to identify a proper \textit{directed acyclic graph} (DAG) $\mathcal{G}(\mathbf{V},\mathbf{E})$ among event types $\mathbf{V}$ such that for a type-$v$ event, the intensity $\lambda_v(t)$ only depends on the event from its cause $\mathbf{Pa}_{v}$ where $\mathbf{Pa}_{v}$ is the parent set of $v$ according to the edge set $\mathbf{E}$, i.e., $\{v'\to v|v'\in \mathbf{Pa}_v\} \subseteq \mathbf{E}$. 
Note that a DAG constraint for the instantaneous causal relation is necessary due to the lack of temporal precedence but it can be easily relaxed by taking lagged and instantaneous relations separately which we omit here for simplicity.

To learn causal structure, we parameterize the causal influence into the reproduction function $\phi_{v',v}(t)=\alpha_{v',v}\kappa(t)$ where $\alpha_{v',v}$ denotes the causal strength of causal relation $v'\to v$ and $\kappa(t)$ characterizes the time-decay of the causal influence which is usually set as an exponential form $\kappa(t)=\exp(\beta t)$ for $t\ge 0$ with the hyper-parameter $\beta$. That is, the causal relationship can be encoded by the impact function such that for any pairs $v',v\in \mathbf{V}$ if $\phi_{v',v}(t)=0$ we have $v'\to v \not\in \mathbf{E}$.

Thus, one of the challenges of learning causal structure is to constrain the sparsity of the reproduction function. To this end, we devise the objective function with sparsity constraint using the $\ell_0$ norm under the likelihood framework.

Given a collection of discrete-time sequences $\mathbf{X}$, the log-likelihood of parameters $\Theta=\{\mathbf{A}=[\alpha_{v',v}]\in \mathbb{R}^{|\mathbf{V}|\times |\mathbf{V}|},\mathbf{\mu}=[\mu_v]\in \mathbb{R}^{|\mathbf{V}|}\}$ of SHP can be expressed as follows:
\begin{equation}\label{eq:likelihood}
\begin{aligned}
 & \mathcal{L} (\mathcal{G} ,\Theta ;\mathbf{X} )\\
= & \sum\limits _{v\in \mathbf{V}}\sum\limits ^{K}_{k=1} [-\lambda _{v} (k\Delta )\Delta +X_{v,k}\log (\lambda _{v} (k\Delta ))]\\
 & +\underbrace{\sum\limits _{v\in \mathbf{V}}\sum\limits ^{K}_{k=1} [-\log (X_{v,k} !)+X_{v,k}\log (\Delta )]}_{\text{Const.}}\\
= & \sum\limits _{v\in \mathbf{V}}\sum\limits ^{K}_{k=1}\left[ -\left( \mu _{v} +\sum _{v'\in \mathbf{V}}\sum ^{k}_{i=1} \phi _{v',v} ((k-i)\Delta )X_{v,i}\right) \Delta \right. \\
 & \left. +X_{v,k}\log\left( \mu _{v} +\sum _{v'\in \mathbf{V}}\sum ^{k}_{i=1} \phi _{v',v} ((k-i)\Delta )X_{v,i}\right)\right] +\text{Const.}
\end{aligned}
\end{equation}
Without further constraint, the log-likelihood function will tend to produce excessive redundant causal edges. Thus, we further penalize the model with $\ell_0$ norm to enforce the sparsity and we obtain the objective function as follows:
\begin{equation} \label{eq:obj}
    \mathcal{L}_p(\mathcal{G} ,\Theta ;\mathbf{X} )=  \mathcal{L}(\mathcal{G} ,\Theta ;\mathbf{X} )+\alpha_{S}\|\mathbf{A}\|_0,
\end{equation}
where $\alpha_{S}$ controls the strength of $\ell_0$ norm. Although there are different forms of sparse regularization for the constraint, e.g., \cite{xu2016learning} proposed to use the $\ell_1$ and $\ell_2$ norm, the $\ell_0$ can provide better sparsity performance and it is easy to be extended, for example, considering the directed acyclic constraint \cite{tsamardinos2006max}, as many causal structures in real-world are often directed acyclic.

\subsection{A Minorization-Maximization-based Algorithm for Learning Causal Structure}

However, estimation of the parameters by maximizing the likelihood in Eq. \ref{eq:obj} has two obstacles. First, the likelihood of the point processes model is known as flat making the optimization unstable and computationally intensive \cite{veen2008estimation}. Second, learning DAGs from observational data is a combinatorial problem and, without any assumption, it has been shown to be
NP-hard \cite{chickering2004large}. To tackle the issues above, following \cite{lewis2011nonparametric,chickering2002optimal}, we propose to use minorization-maximization (MM) based optimization \cite{hunter2004tutorial} with a two-step causal structure learning algorithm. 

For the first step, the parameters are estimated given a fixed causal graph using the MM algorithm which leverages the additive structure in Eq. \ref{eq:likelihood} to apply Jensen's inequality similar to the EM algorithm and obtain the following lower bound:
\begin{equation}\label{eq:upper_bound}
\begin{aligned}
 & Q(\Theta |\Theta ^{(j)} )\\
= & \sum\limits _{v\in \mathbf{V}}\sum\limits ^{K}_{k=1}\left[ -\left( \mu _{v} +\sum _{v'\in \mathbf{V}}\sum ^{k}_{i=1} \phi _{v',v} ((k-i)\Delta )X_{v',i}\right) \Delta \right. \\
 & +X_{v,k}\left({\displaystyle q^{\mu }_{v,k}}\log\left(\frac{\mu _{v}}{{\displaystyle q^{\mu }_{v,k}}}\right)\right. \\
 & \left. \left. +\sum _{v'\in \mathbf{V}}\sum ^{k}_{i=1}{\displaystyle q^{\alpha }_{v,k} (v',i)}\log\left(\frac{\phi _{v',v} ((k-i)\Delta )X_{v',i}}{{\displaystyle q^{\alpha }_{v,k} (v',i)}}\right)\right)\right]
\end{aligned}
\end{equation}
where $q^{\mu }_{v,k} =\frac{\mu ^{(j)}_{v}}{\lambda ^{(j)}_{v} (k\Delta )}$ and $q^{\alpha }_{v,k}( v',i) =\frac{\phi ^{( j)}_{v',v} ((k-i)\Delta )X_{v',i}}{\lambda ^{(j)}_{v} (k\Delta )}$. $\lambda ^{(j)}_{v} (k\Delta )$ is the conditional intensity function with parameters $\Theta^{(j)}$ in the $j$-th iteration. Then, by setting $\frac{\partial Q(\Theta |\Theta ^{(j)} )}{\partial \mu _{v}} =0$, $\frac{\partial Q(\Theta |\Theta ^{(j)} )}{\partial \alpha _{v',v}} =0$, we obtain the close-form iteration formulas:
\begin{equation}\label{eq:iter}
\begin{aligned}
\mu ^{(j+1)}_{v} & =\frac{\sum\nolimits ^{K}_{k=1} X_{v,k} q^{\mu }_{v,k}}{K\Delta }\\
\alpha ^{(j+1)}_{v',v} & =\begin{cases}
\frac{\sum\nolimits ^{K}_{k=1}\sum ^{k}_{i=1} q^{\alpha }_{v,k} (v',i)X_{v,k}}{\sum\nolimits ^{K}_{k=1}\sum ^{k}_{i=1} \kappa ((k-i)\Delta )X_{v',i} \Delta } & v'\neq v\\
\frac{\sum\nolimits ^{K}_{k=1}\sum\nolimits ^{k-1}_{i=1} q^{\alpha }_{v,k} (v',i)X_{v,k}}{\sum\nolimits ^{K}_{k=1}\sum\nolimits ^{k-1}_{i=1} \kappa ((k-i)\Delta )X_{v',i} \Delta } & v'=v
\end{cases}
\end{aligned}
\end{equation}

\begin{algorithm}[tb]
	\caption{Learning causal structure using SHP}
	\label{alg:hc}
        \textbf{Input}: Data set $\mathbf{X}$\\
        \textbf{Output}: $G^*,\Theta^*$
	\begin{algorithmic}[1] 
		\STATE $G' \gets empty~graph$, $\mathcal{L}^*_p\gets -\infty$
		\WHILE {$\mathcal{L}^*_p(G^*,\Theta^*;\mathbf{X})< \mathcal{L}'_p(G',\Theta';\mathbf{X})$}
		\STATE $G^*,\Theta^*\gets G',\Theta'$ with largest $\mathcal{L}'_p(G',\Theta';\mathbf{X})$
		\FOR {every $G' \in \mathcal{V}(G^*)$}
		    \STATE Update $\Theta'$ via iteration in Eq. \ref{eq:iter}
		    \STATE Record score $\mathcal{L}'_p(G',\Theta';\mathbf{X})$
		\ENDFOR
		\ENDWHILE
		\STATE \textbf{return} $G^*,\Theta^*$
	\end{algorithmic}
\end{algorithm}

With the MM algorithm, we then search the causal structure by using a Hill-Climbing-based algorithm as shown in Algorithm \ref{alg:hc}. It mainly consists of two phases. First, we perform a structure searching scheme by taking one step adding, deleting, and reversing the graph $G^*$ in the last iteration, i.e., in Line 4, $\mathcal{V}(G^*)$ represents a collection of the one-step modified graph of $\mathcal{V}(G^*)$. Furthermore, the acyclic constraint is implemented by eliminating all cyclic causal graphs in $\mathcal{V}(G^*)$. Second, by fixing the graph $G'$, we optimize the log-likelihood using the MM algorithm in Line 5. Iterating the two steps above until the likelihood no longer increases. 
\section{Identifiability}\label{sec:identifiability}

In this section, we aim to answer the question that whether we can identify the causal relationship under the existence
of the instantaneous effect. In answering this question, one will need to explore the property of Hawkes processes in discrete-time. Based on the discrete-time likelihood in Eq. \ref{eq:likelihood}, each interval is modeled by conditional Poisson distribution with the linear structure in Eq. \ref{eq:discrete_intensity_continue_instant}.  As such, one may alternatively represent the relation by integer-valued autoregressive (INAR) processes according to \cite{kirchner2016hawkes} for analyzing the identification of structural Hawkes Processes. Furthermore, we assume that the causal sufficiency holds, i.e., all relevant variables have been observed \cite{spirtes2000causation}.
Following \cite{kirchner2016hawkes}, the INAR($\infty$) processes can be defined as follows:
\begin{definition}[INAR($\infty$)]
For $\theta_k \geq 0,k\in \mathbb{N}_0$, let $\epsilon_t \stackrel{i.i.d.}{\sim } \operatorname{Pois}(\theta_0),t\in \mathbb{N}$, and $\xi_{i}^{(t,k)}\sim \operatorname{Pois}(\theta_k)$. An Integer-valued autoregressive time series of infinite order (INAR($\infty$)) process ${X_t}, t\in \mathbb{N}$ is defined by
\begin{equation}
    X_t=\sum_{k=1}^{\infty}\theta_k \circ X_{t-k} + \epsilon_t
\end{equation}
where $\circ$ is a reproduction operator given by $\theta_k \circ X_{t-k} \equiv \sum_{i=1}^{X_{t-k}}\xi_{i}^{(t,k)}$ with $\xi_{i}^{(t,k)}$ be a sequence of i.i.d. non-negative integer-valued random variables that depends on the reproduction coefficients $\theta_k$, $\{\epsilon_t\}_{t\in \mathbb{N}}$ is an i.i.d. integer-valued immigration sequence that are independent of $\{\xi_{i}^{(t,k)}\}$, and $X_{t-k}$ is independent of $\epsilon_t$ for all $k$.
\end{definition}
In general, INAR is a discrete analogy of the continuous autoregressive model. Note that the distribution of the independent variables in INAR could be different and different choices of the distribution would lead to different INAR models \cite{guerrero2022integer}. Here, we use the Poisson choice to adopt the conditional Poisson distribution in Eq. \ref{eq:likelihood}, and it has been shown that such INAR($\infty$) model will converge to the continuous-time Hawkes processes as the time interval $\Delta\to 0$:
\begin{theorem}[\cite{kirchner2016hawkes}]\label{thm:INAR_eq}
Let $N$ be a Hawkes process with immigration intensity $\mu$ and let $\phi:\mathbb{R}\to \mathbb{R}_0^+$ be a reproduction intensity that is piecewise continuous with $\phi(t)=0,t\leq 0$ and $\int \phi(t)dt<1$. For $\Delta\in (0,\delta)$, let $\left( X_t^{(\Delta)} \right)$ be an INAR($\infty$) sequence with immigration parameter $\Delta \mu$ and reproduction coefficients $\Delta \phi(k\Delta),k\in \mathbb{N}$. From the sequences ${(X_t^{(\Delta)})}_{\Delta \in (0,\delta)}$, we define a family of point processes by
\begin{equation}
N^{(\Delta)}(A)\coloneq\sum_{k: k \Delta \in A} X_{k}^{(\Delta)}, \quad A \in \mathcal{B}, \Delta \in(0, \delta),
\end{equation}
where $\mathcal{B} \coloneq \mathcal{B}(\mathbb{R})$ is the Borel set in $\mathbb{R}$. Then, we have that
\begin{equation}
N^{(\Delta)} \stackrel{\mathrm{w}}{\longrightarrow} N \quad \text { for } \Delta \rightarrow 0 .
\end{equation}
\end{theorem}
Theorem \ref{thm:INAR_eq} implies that the properties of INAR can be utilized for analyzing the Hawkes processes in discrete-time. Intuitively, the reproduction function $\phi$ in the discrete-time Hawkes processes can be represented by a series of reproduction coefficients $\theta_k$ for each time period, and the immigration intensity $\mu$ in the discrete-time Hawkes processes can be represented by the immigration parameters of $\epsilon_t$.

Specifically, given the property of INAR processes, the analysis of the identifiability of structural Hawkes processes can be typically performed in two folds---the identifiability of the temporal structural Hawkes processes and the instantaneous structural Hawkes processes \cite{hyvarinen2010estimation}. For the former, the temporal resolution of the measurement that is high enough to capture the former and latter relationship of the events, the identifiability can be derived by local independence, which has been well explored by \cite{mogensen2020markov}. For example, in Fig. \ref{fig:toy_example}, the causal relationship $v_1\to v_3$ can be simply identified by the independence. Thus, in this work, we are more interested in the instantaneous structural Hawkes processes---the measurements have lower time resolution such that the causal influences are instantaneous. In such a case, one can use a model in which the influences are instantaneous, leading to Bayesian networks (BNs) or structural equation models (SEMs), e.g., the causal relationship $v_1\to v_2$ in Fig. \ref{fig:toy_example} belong to this class. 

Thus, to analyze the identification of SHP, based on the INAR model, we consider the instantaneous causal structure in the structural Hawkes process:
\begin{definition}[Instantaneous causal structure in structural Hawkes process]\label{def:insem}
Let $\epsilon_{v,t} \stackrel{i.i.d.}{\sim } \operatorname{Pois}(\mu_{v})$, and $\xi_{i}^{(v',v)}\sim \operatorname{Pois}(\alpha_{v',v})$.
The instantaneous causal structure in the structural Hawkes process consists of a set of equations of the form
\begin{equation}
    X_{v,t}=\sum_{v'\in \mathbf{V}} \alpha_{v', v}\circ X_{v',t} +\epsilon_{v},\quad v\in \mathbf{V},
\end{equation}
where $\alpha_{v', v}>0$ for $v'\in \mathbf{Pa}_v$ and $\alpha_{v', v}=0$ for $v'\not \in \mathbf{Pa}_v$, with $\alpha_{v', v}\circ X_{v',t}\equiv \sum_{i=1}^{X_{v',t}}\xi_i^{(v,v')}$, and $\mu_v>0$ if $v$ is a root variable i.e., the variable whose parent set is empty $\mathbf{Pa}_v=\phi$. The random variables $X_{v',t}$, $\xi_i^{(v,v')}$, and $\epsilon_v$ are independent of each other.
\end{definition}
The identifiability of the instantaneous causal structure, however, is much more difficult compared with the temporal causal relationship and most of them suffer from lack of identifiability. For example, without any constraint, the instantaneous causal structure may not be identified due to the Markov equivalence class \cite{pearl2009causality} in which all graphs in the equivalence class encode the same conditional independence. In such a case, different model with different causal graph and parameter, $M_1=(\mathcal{G}_1,\Theta_1)$, $M_2=(\mathcal{G}_2,\Theta_2)$ will produce the same distribution. Such non-identifiability also exists even considering an additional constraint, e.g., the linear Gaussian SEM is also non-identifiable \cite{spirtes2000causation}. 

Thus, an essential goal of this work is to investigate the identifiability of the instantaneous causal structure, i.e., whether there exists another causal structure that entails the same distribution to the underlying causal model. Such identifiability basically boils down to bivariate cases that whether there exists a backward model that is distribution equivalent to the causal direction. It can be easily extended to multivariate cases by incorporating the conditional independent constraint in the causal structure with the causal faithfulness assumption, i.e., the causal graph faithfully displays every dependency \cite{pearl1988probabilistic}.
Specifically, let $X$ be the cause variable and $Y$ be the effect variable, and surprisingly, the following theorem shows that the bivariate instantaneous causal pair is indeed identifiable:
\begin{theorem}\label{thm:identifiability}
Let $X\to Y$ be the correct causal direction that follows
\begin{equation}\label{eq:identifiability_model}
	\begin{aligned}[c]
	Y&=\sum\nolimits_{i=1}^{X}\xi_i + \epsilon,&\quad X, \xi_i, \textrm{ and }\epsilon \textrm{~are independent},\\
	\end{aligned}
\end{equation}
where $\xi_i\sim \operatorname{Pois}(\alpha_{X,Y})$, $\epsilon\sim \operatorname{Pois}(\mu_Y)$, $X\sim \operatorname{Pois}(\mu_X)$.
Then, there does not exist a backward model that admits the following equation:
\begin{equation}\label{eq:identifiability_backward_model}
	\begin{aligned}[c]
	 X&=\sum\nolimits_{i=1}^{Y}\hat{\xi}_i + \hat{\epsilon}, &\quad Y,\hat{\xi}_i, \textrm{ and }\hat{\epsilon } \textrm{~are independent},
	\end{aligned}
\end{equation}
where $\hat{\xi}_i\sim \operatorname{Pois}(\hat{\alpha}_{Y,X})$, $\hat{\epsilon}\sim \operatorname{Pois}(\hat{\mu}_X)$, $Y\sim \operatorname{Pois}(\hat{\mu}_Y)$.
\end{theorem}
To better understand the identifiability, we provide two alternative proofs for this theorem. The details of the proofs are given in the supplementary material. The first proof show that it is impossible to have a backward model that the probability distribution of the two directions is equivalence. While the second proof shows that the distribution of $Y$ must not admit the Poisson distribution and therefore the distribution is not equivalent. We also provide an empirical study of the identifiability for the bivariate causal pair in Section \ref{sec:experiments}. Finally, we generalize the result of Theorem \ref{thm:identifiability} to the multivariate causal structure and show that the multivariate instantaneous causal structure is also identifiable:
\begin{theorem}
With the causal faithfulness assumption and causal sufficiency assumption, the multivariate instantaneous causal structure is identifiable.
\end{theorem}
The main idea behind the proof is that for any causal structure $(\mathcal{G},\Theta)$. there does not exist another causal structure $(\hat{\mathcal{G}},\hat{\Theta})$ that is distribution equivalent. Specifically, with the causal faithfulness and causal sufficiency assumption, the Markov equivalent class is identified. Since all Markov equivalent classes share the same skeleton, we only need to show that each causal pair will not admit another causal structure that has the revered causal direction while having the same likelihood (i.e., distribution equivalent).
\section{Experiments} \label{sec:experiments}

In this section, we test the proposed SHP and the baselines on both synthetic and real-world data. The baseline methods include ADM4 \cite{zhou2013learningsocial}, NPHC \cite{achab2017uncovering}, MLE\_SGL \cite{xu2016learning}, and PCMCI Plus \cite{runge2019detecting}. We further develop SHP\_NH, an ablation study of SHP that removes the hill-climb based searching scheme and uses a threshold to determine the causal structure following the work of ADM4 and MLE\_SGL. In all following experiments, Recall, Precision, and F1 are used as the evaluation metrics. The results of recall and precision are provided in the supplementary material.

\subsection{Synthetic Experiments}

\begin{figure}
    \centering
    \includegraphics[width=0.4\textwidth]{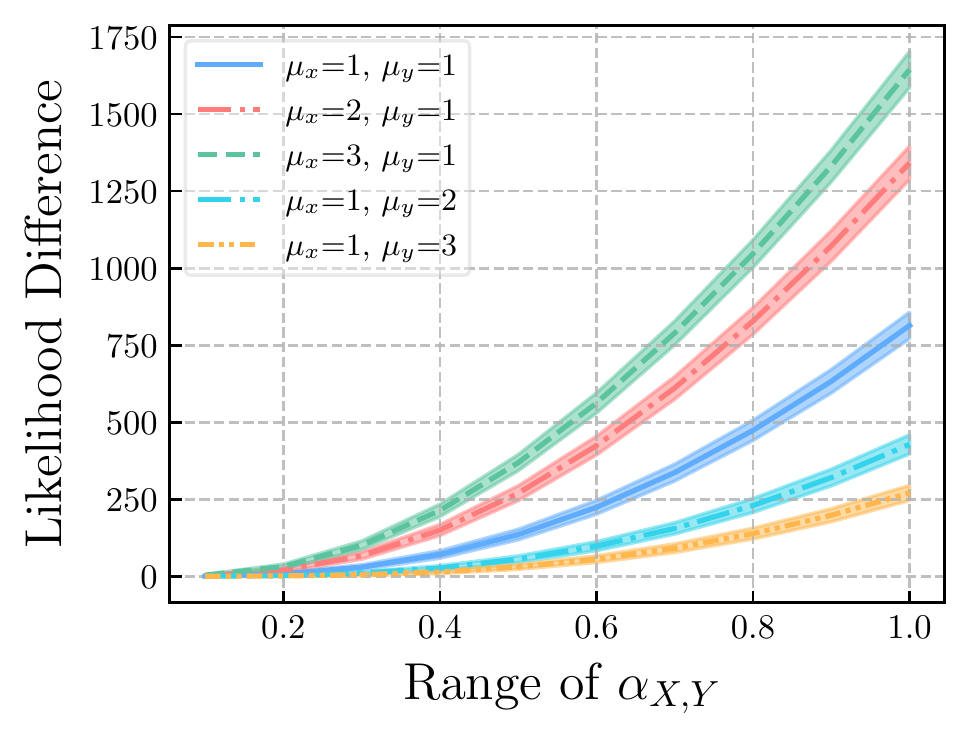}
    \caption{The likelihood difference between two causal pairs in Theorem \ref{thm:identifiability} with different causal influence and immigration intensity.}
    \label{fig:insems_theta}
\end{figure}

\begin{figure*}[t]
	\centering
	\subfigure[Sensitivity to Time Interval]{
	\includegraphics[width=0.33\textwidth]{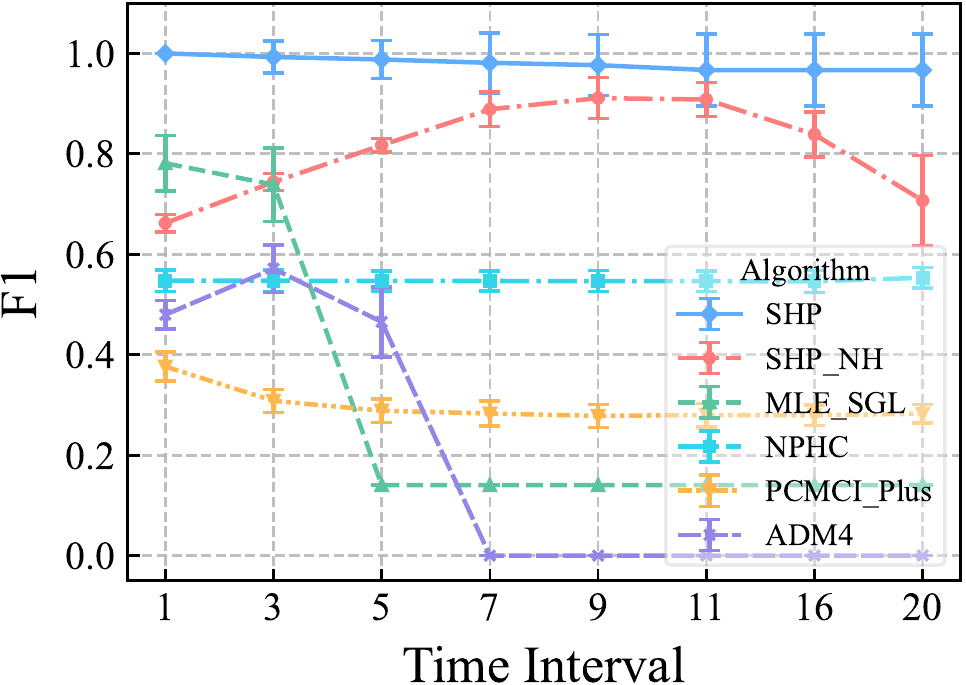}
	\label{fig:sensitivity:interval}
}
	\subfigure[Sensitivity to Range of $\alpha$]{
		\includegraphics[width=0.31\textwidth]{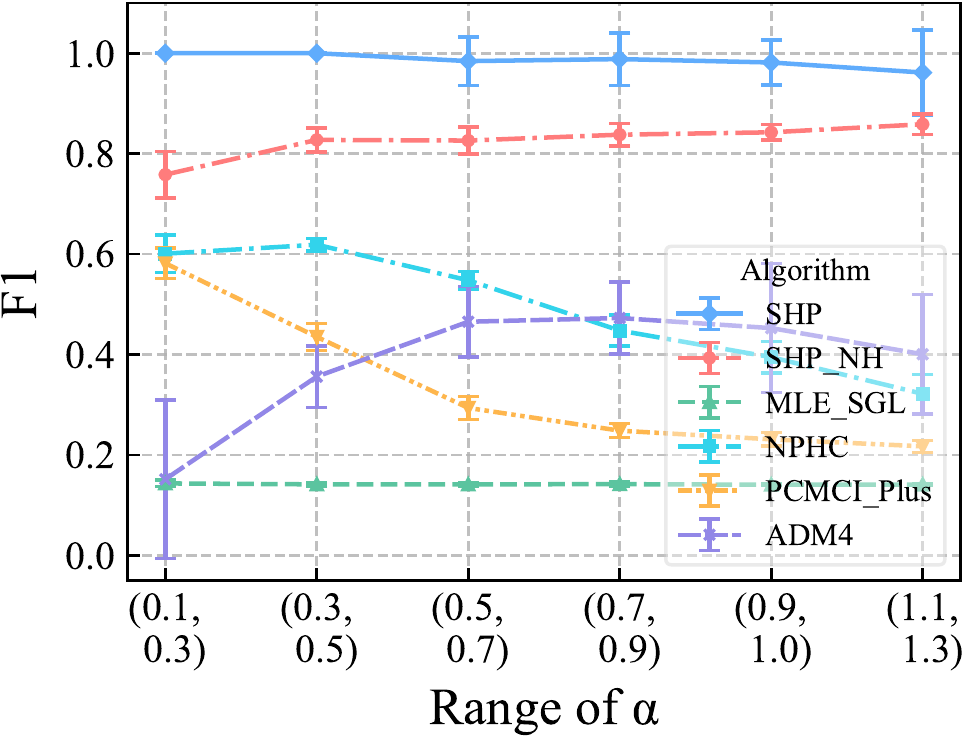}
		\label{fig:sensitivity:alpha}
	}
	\subfigure[Sensitivity to Range of $\mu$]{
	\includegraphics[width=0.31\textwidth]{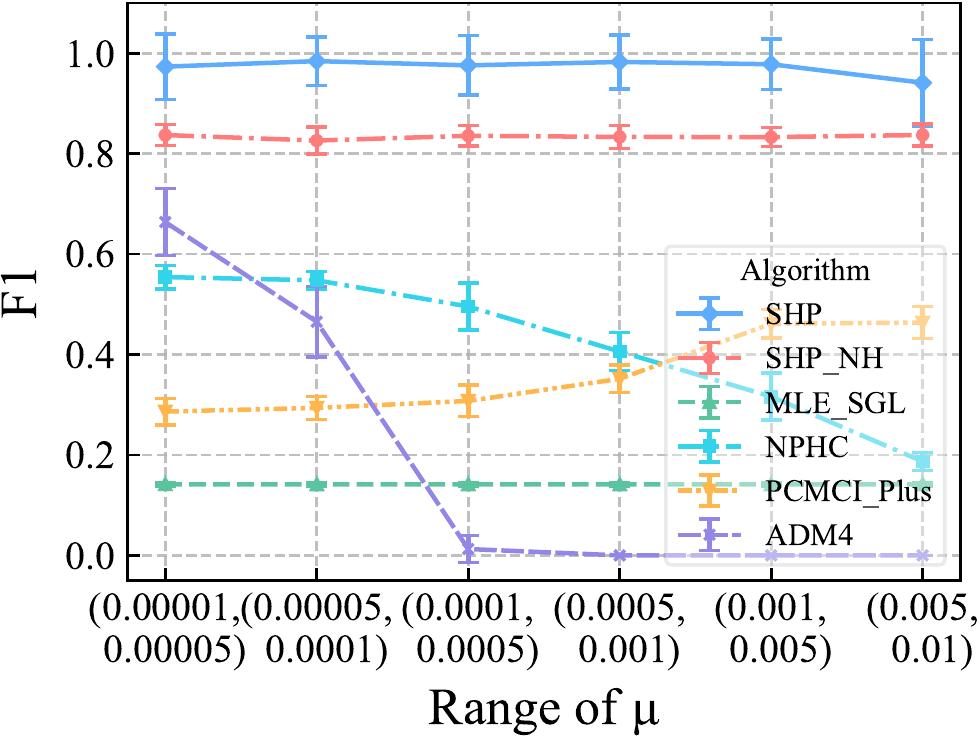}
	\label{fig:sensitivity:mu}
}
	\subfigure[Sensitivity to Sample Size]{
	\includegraphics[width=0.32\textwidth]{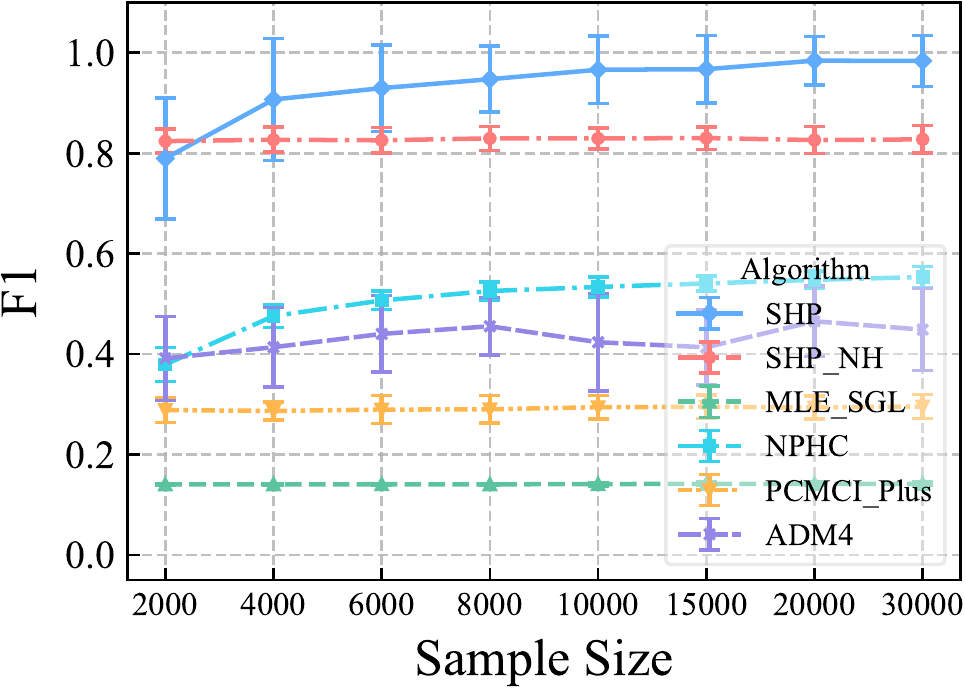}
	\label{fig:sensitivity:sample}
}
	\subfigure[Sensitivity to Num. of Event Types]{
	\includegraphics[width=0.32\textwidth]{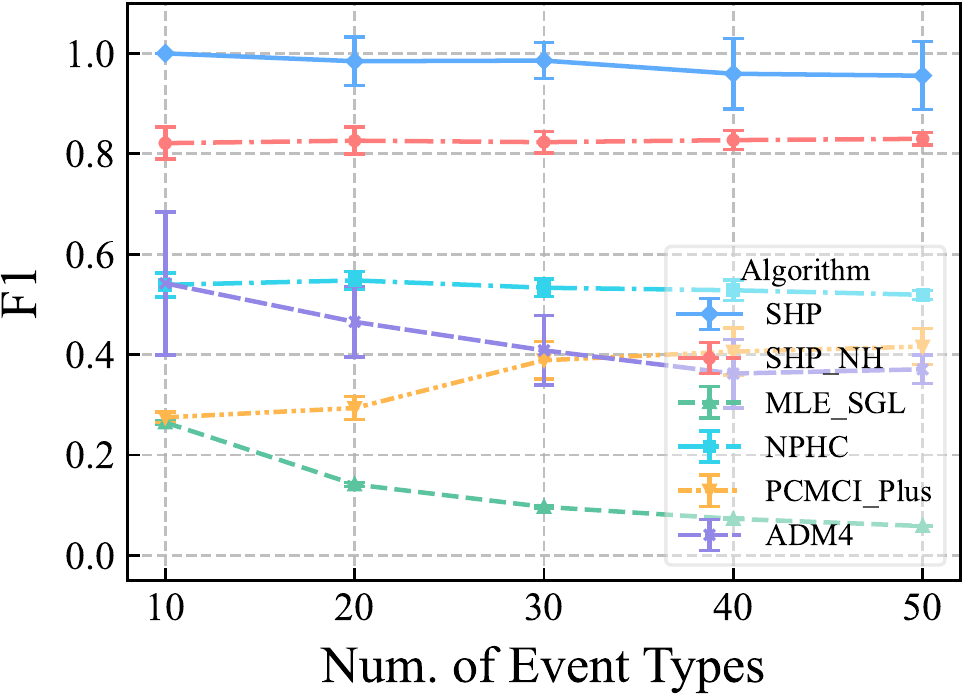}
	\label{fig:sensitivity:num_type}
}
	\subfigure[Sensitivity to Avg. Indegree]{
	\includegraphics[width=0.32\textwidth]{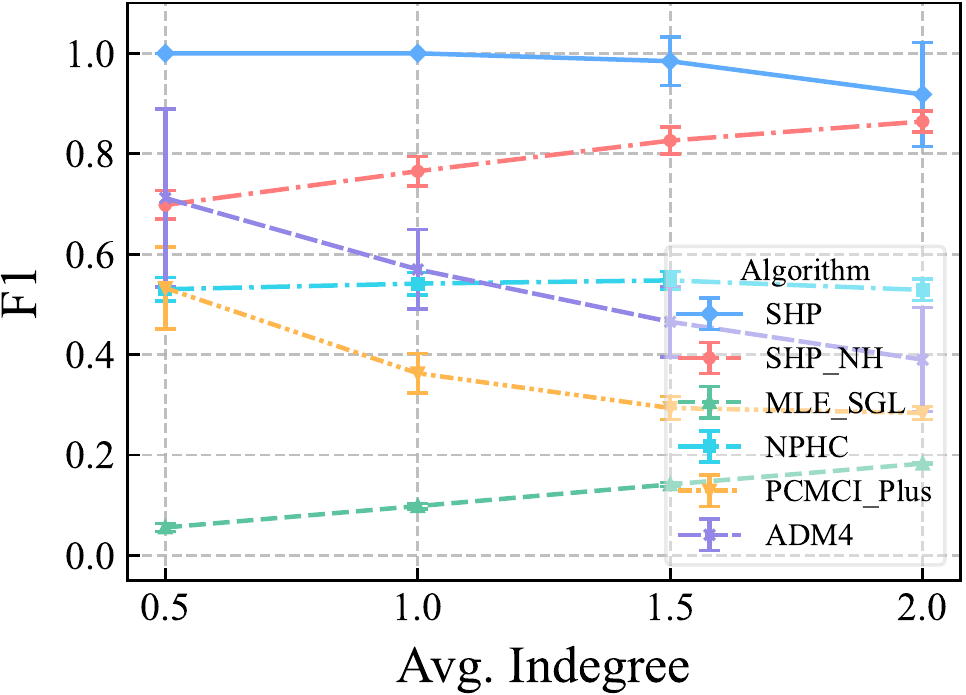}
	\label{fig:sensitivity:degree}
}
	\caption{F1 in the Sensitivity Experiments}	
	\label{fig:sensitivity}
\end{figure*}

In this part, we design extensive control experiments using synthetic data to test the correctness of the theory and the sensitivity of sample size, length of time interval, number of event types, and different ranges of $\alpha$ and $\mu$. In the sensitivity experiment, we synthesize data with fixed parameters while traversing the target parameter as shown in Fig. \ref{fig:sensitivity}. The default settings are listed below, sample size=20000, time interval=5, number of event types=20, range of $\alpha\in [0.3,0.5]$, range of $\mu\in [0.0005,0.0001]$. All experimental results are averaged over 100 randomly generated causal structures.

The generating process proceeds as follows: 1) randomly generate a directed causal graph $\mathcal{G}$ with a certain average in degree; 2) generate the events according to randomly generated parameters $\alpha_{v',v},\mu_v$ from $\mathcal{G}$, and 3) aggregate the number of counts at each interval for each event according to the length of time interval to synthesize the discrete-time process.

\paragraph{Two-variable case.}
We first conduct a simple two-variable experiment to verify the identifiability in Theorem \ref{thm:identifiability}, by computing the likelihood difference between the causal direction and the reversed direction on a simulated causal pair with different causal influence $\alpha_{X,Y}$. That is, we simulated data using the model $Y_t=\alpha_{X,Y}\circ X_t+\epsilon_t$ with $X_t\sim \operatorname{Pois}(\mu_x)$ and $\epsilon_t\sim  \operatorname{Pois}(\mu_y)$. Each experiment is conducted 100 times with random causal pairs. As shown in Fig. \ref{fig:insems_theta}, we can see that the likelihood difference is always greater than zero which means that it is always identifiable unless the degenerate cases and it verifies the correctness of Theorem \ref{thm:identifiability}. In addition, as the causal influence $\alpha_{X,Y}$ decreases, the likelihood difference also decreases and tends to zero. This is reasonable as the causal influence becomes zero, and the two variables will also become independent making the model non-identifiable. Moreover, the level of immigration intensity also would affect the likelihood difference, which is reasonable as the lower the $\mu_x$, the weaker the causal relation. Similarly, the higher the $\mu_y$, the stronger the noise making  the causal relation weaker. 
The more general multivariate case experiments will be implied in the following sensitivity analysis.

\paragraph{Sensitivity analysis.}
As shown in Fig. \ref{fig:sensitivity}, we conduct six different control experiments for SHP. In general, our proposed SHP method outperforms all the baseline methods in all six control experiments. 

In the control experiments of time interval given in Fig. \ref{fig:sensitivity:interval}, as the time interval controls the  temporal resolution of the measurement sequence, the larger the time interval, the lower the temporal resolution, and the more instantaneous causal influences will occur leading to the decrease or insensitive of performance of the baseline methods. In the contract, SHP keeps giving the best results at all intervals. We also notice that the performance will decrease as the time interval increase because it reduces the sample size and the sequence become less informative. In addition, the performance of SHP\_NH stresses the effectiveness of the searching algorithm which has lower precision and stability than SHP but also outperforms the baseline methods if the event sequences contain sufficient information on the instantaneous effect, which also verifies the ability to capture the instantaneous effect in SHP.

For the controlled experiments of causal strength and immigration intensity in Fig. \ref{fig:sensitivity:alpha} and Fig. \ref{fig:sensitivity:mu}, we can see that a reasonable causal strength and immigration intensity are required for the baseline methods while SHP is insensitive to both of them which shows the robustness of SHP.

In the sample size controlled experiments given in Fig. \ref{fig:sensitivity:sample}, all methods are robust to the sample size, and in particular, a 1000 sample size is enough to produce a reasonable performance for SHP, which demonstrates the practicality of SHP. 

For the causal structure controlled experiments given in Fig. \ref{fig:sensitivity:num_type} and Fig. \ref{fig:sensitivity:degree}, SHP performs well in both experiments, and the average indegree is also important as it increases the performance of most methods decreases but SHP has the smallest decline.

\subsection{Real World Experiments}

\begin{figure}[t]
    \centering
    \includegraphics[width=0.4\textwidth]{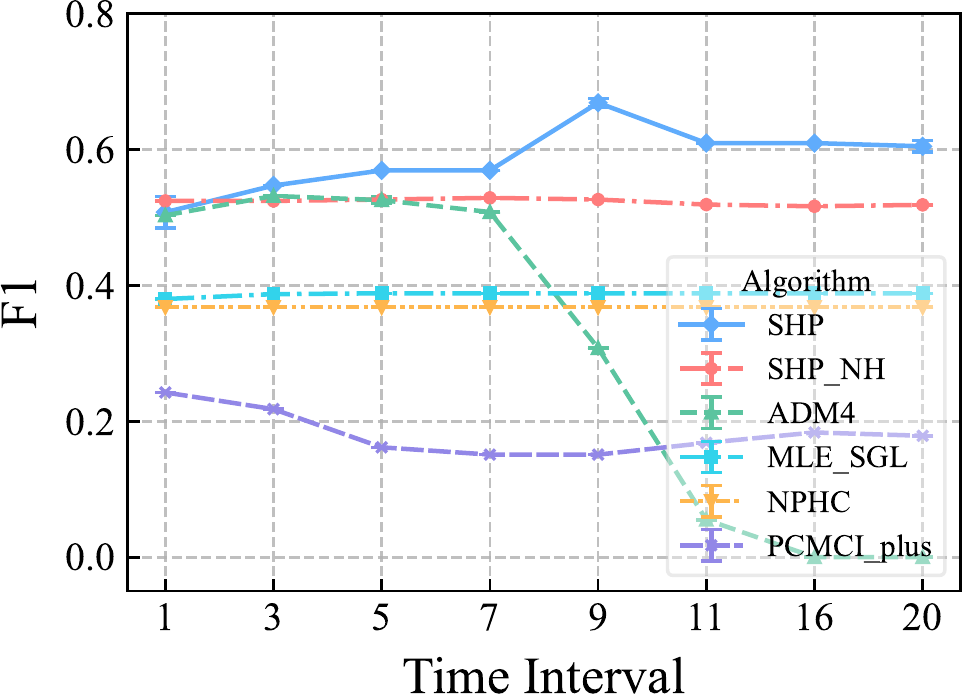}
    \caption{Real-world experiment on different temporal resolutions.}
    \label{fig:real_world_time_interval}
\end{figure}

We also test the proposed SHP on a very challenging real-world dataset\footnote{\url{https://competition.huaweicloud.com/informations/mobile/1000041487/dataset}} from real telecommunication networks. The dataset records eight months of alarms that occurred in a real metropolitan cellular network. The alarms are generated by fifty-five devices in the metropolitan cellular network, which consists of eighteen types of alarms. Our goal is to learn the causal structure among the alarm types. Note that the causal impact among alarms is not deterministic but probabilistic in this system, which is similar to our model setting. All experiments have been conducted with different random seeds and the results are significant according to Wilcoxon signed-rank test. 
In addition, the ground truths of the causal relationships among alarm types are provided by domain experts.

As shown in Fig. \ref{fig:real_world_time_interval}, we conduct the real-world experiments by manually setting different temporal resolutions of the observed sequence (i.e., from one second to nine seconds) and the original temporal resolution is one second. Interestingly, as the time interval increases, the performance of our method also increases instead of decreasing, while the baseline methods all decrease. The reason is that in the real-world scenario, there exists a communication latency in the logging system such that the recorded timestamp might be not fully accurate, and this error can be mitigated by decreasing the temporal resolution. In the contrast, after decreasing the temporal resolution, other methods fail to capture the causal relationship under the instantaneous effect. This verifies the effectiveness of SHP and stresses the importance of the instantaneous effect.

\section{Conclusion}
In this work, we study how to model and leverage the instantaneous effect for learning causal structure in discrete-time event sequences. We propose structural Hawkes processes that leverage the instantaneous effects and a practical algorithm for learning the causal structure among event types. Theoretical results show that the instantaneous causal structure in structural Hawkes processes is indeed identifiable. To the best of our knowledge, this is the first causal structure learning method for event sequences with instantaneous effects. The success of SHP not only provides an effective solution for learning causal structure from real-world event sequences but also shows a promising direction for causal discovery from the discrete-time event sequences. In the future, we plan to extend our work to a general point process with a more general intensity function.

\section*{Acknowledgements}
This research was supported in part by National Key R\&D Program of China (2021ZD0111501), National Science Fund for Excellent Young Scholars (62122022), Natural Science Foundation of China (61876043, 61976052), the major key project of PCL (PCL2021A12). We appreciate the comments from anonymous reviewers, which greatly helped to improve the paper.
\bibliographystyle{named}
\bibliography{ijcai23}

\newpage
\onecolumn
\appendix

\newtheorem{innercustomthm}{Theorem}
\newenvironment{customthm}[1]
  {\renewcommand\theinnercustomthm{#1}\innercustomthm}
  {\endinnercustomthm}

\numberwithin{equation}{section}

\setcounter{section}{0}
\section*{Supplementary Material}
In this supplementary material, we provide the derivation of minorization-maximization algorithm in Section \ref{asec:em}, the proofs of Theorem 2 in Section \ref{asec:proof2}, the proof of Theorem 3 in Section \ref{asec:proof3}, and the additional experiments results in Section \ref{asec:exp}.

\section{Derivation of minorization-maximization algorithm}\label{asec:em}
In this section, we provide the detailed derivation of the proposed MM algorithm. Given a set of observations $\mathbf{X}$, the log-likelihood of the causal graph $\mathcal{G}$ and parameters $\Theta$, the log-likelihood can be derived as follows:
\begin{equation}\label{aeq:likelihood}
\begin{aligned}
 & \mathcal{L} (\mathcal{G} ,\Theta ;\mathbf{X} )\\
= & \sum\limits _{v\in \mathbf{V}}\sum\limits ^{K}_{k=1}\log\left[\frac{e^{-\lambda _{v} (k\Delta )\Delta }}{X_{v,k} !} (\lambda _{v} (k\Delta ))\Delta )^{X_{v,k}}\right]\\
= & \sum\limits _{v\in \mathbf{V}}\sum\limits ^{K}_{k=1} [-\lambda _{v} (k\Delta )\Delta +X_{v,k}\log (\lambda _{v} (k\Delta ))]\\
 & +\underbrace{\sum _{v\in \mathbf{V}}\sum\nolimits ^{K}_{k=1} [-\log (X_{v,k} !)+X_{v,k}\log (\Delta )]}_{\coloneq \text{Const}}\\
= & \sum\limits _{v\in \mathbf{V}}\sum\limits ^{K}_{k=1}\left[ -\left( \mu _{v} +\sum _{v'\in \mathbf{V}}\sum ^{k}_{i=1} \phi _{v',v} ((k-i)\Delta )X_{v',i}\right) \Delta \right. \\
 & \left. +X_{v,k}\log\left( \mu _{v} +\sum _{v'\in \mathbf{V}}\sum ^{k}_{i=1} \phi _{v',v} ((k-i)\Delta )X_{v',i}\right)\right] +\text{Const}\\
= & \sum\limits _{v\in \mathbf{V}}\sum\limits ^{K}_{k=1}\left[ -\left( \mu _{v} +\sum _{v'\in \mathbf{V}}\sum ^{k-1}_{i=1} \alpha _{v',v} \kappa ( (k-i)\Delta ) X_{v',i} +\sum _{v'\in \mathbf{V} \backslash v} \alpha _{v',v} \kappa ( 0) X_{v',k}\right) \Delta \right. \\
 & \left. +X_{v,k}\log\left( \mu _{v} +\sum _{v'\in \mathbf{V}}\sum ^{k-1}_{i=1} \alpha _{v',v} \kappa ( (k-i)\Delta ) X_{v',i} +\sum _{v'\in \mathbf{V} \backslash v} \alpha _{v',v} \kappa ( 0) X_{v',k}\right)\right] +\text{Const} ,
\end{aligned}
\end{equation}
where $\displaystyle \phi _{v',v} (t)=\begin{cases}
0 & v'=v\ \text{and} \ t=0\\
\alpha _{v',v} \kappa ( t) & \text{otherwise}
\end{cases}$. By applying the Jensen inequality to $\log(\lambda_v(t))$, we obtain the lower bound of the intensity function:
\begin{equation}\label{aeq:upper_bound}
\begin{aligned}
 & \log( \lambda _{v} (k\Delta ))\\
= & \log\left( \mu _{v} +\sum _{v'\in \mathbf{V}}\sum ^{k-1}_{i=1} \phi _{v',v} ((k-i)\Delta )X_{v',i} +\sum _{v'\in \mathbf{V} \backslash v} \alpha _{v',v} \kappa ( 0) X_{v',k}\right)\\
= & \log\left({\displaystyle q^{\mu }_{v,k}\frac{\mu _{v}}{{\displaystyle q^{\mu }_{v,k}}}} +\sum _{v'\in \mathbf{V}}\sum ^{k-1}_{i=1}{\displaystyle q^{\alpha }_{v,k}( v',i)\frac{\alpha _{v',v} \kappa ( (k-i)\Delta ) X_{v',i}}{{\displaystyle q^{\alpha }_{v,k}( v',i)}}} +{\displaystyle \sum _{v'\in \mathbf{V} \backslash v} q^{\alpha }_{v,k}( v',k)\frac{\alpha _{v',v} \kappa ( 0) X_{v',k}}{{\displaystyle q^{\alpha }_{v,k}( v',k)}}}\right)\\
\geqslant  & {\displaystyle q^{\mu }_{v,k}}\log\left(\frac{\mu _{v}}{{\displaystyle q^{\mu }_{v,k}}}\right) +{\displaystyle \sum _{v'\in \mathbf{V}}\sum ^{k-1}_{i=1} q^{\alpha }_{v,k}( v',i)}\log\left({\displaystyle \frac{\alpha _{v',v} \kappa ( (k-i)\Delta ) X_{v',i}}{{\displaystyle q^{\alpha }_{v,k}( v',i)}}}\right) +{\displaystyle \sum _{v'\in \mathbf{V} \backslash v} q^{\alpha }_{v,k}( v',k)}\log\left({\displaystyle \frac{\alpha _{v',v} \kappa ( 0) X_{v',k}}{{\displaystyle q^{\alpha }_{v,k}( v',k)}}}\right),
\end{aligned}
\end{equation}
where ${\displaystyle q^{\mu }_{v,k} =\frac{\mu ^{(j)}_{v}}{\lambda ^{(j)}_{v} (k\Delta )}}$ and ${\displaystyle q^{\alpha }_{v,k}( v',i) =\frac{\phi ^{( j)}_{v',v} ((k-i)\Delta )X_{v',i}}{\lambda ^{(j)}_{v} (k\Delta )}}$, and $\lambda ^{(j)}_{v} (k\Delta )$ is the conditional intensity function in the $j$-th iteration.
By subsisting Eq. \ref{aeq:upper_bound} into \ref{aeq:likelihood}, we obtain the objective function in our work:
\begin{equation}
\begin{aligned}
 & Q(\Theta |\Theta ^{( j)} )\\
= & \sum\limits _{v\in \mathbf{V}}\sum\limits ^{K}_{k=1}\left[ -\left( \mu _{v} +\sum _{v'\in \mathbf{V}}\sum ^{k}_{i=1} \phi _{v',v} ((k-i)\Delta )X_{v',i}\right) \Delta \right. \\
+ & \left. X_{v,k}\left({\displaystyle q^{\mu }_{v,k}}\log\left(\frac{\mu _{v}}{{\displaystyle q^{\mu }_{v,k}}}\right) +\sum _{v'\in \mathbf{V}}\sum ^{k}_{i=1}{\displaystyle q^{\alpha }_{v,k}( v',i)}\log\left(\frac{\phi _{v',v} ((k-i)\Delta )X_{v',i}}{{\displaystyle q^{\alpha }_{v,k}( v',i)}}\right)\right)\right]\\
= & \sum\limits _{v\in \mathbf{V}}\sum\limits ^{K}_{k=1}\left[ -\left( \mu _{v} +\sum _{v'\in \mathbf{V}}\sum ^{k-1}_{i=1} \alpha _{v',v} \kappa ( (k-i)\Delta ) X_{v',i} +\sum _{v'\in \mathbf{V} \backslash v} \alpha _{v',v} \kappa ( 0) X_{v',k}\right) \Delta \right. \\
+ & \left. X_{v,k}{\displaystyle q^{\mu }_{v,k}}\log\left(\frac{\mu _{v}}{{\displaystyle q^{\mu }_{v,k}}}\right) +{\displaystyle X_{v,k}}\sum _{v'\in \mathbf{V}}\sum ^{k-1}_{i=1}{\displaystyle q^{\alpha }_{v,k}( v',i)}\log\left(\frac{\alpha _{v',v} \kappa ( (k-i)\Delta ) X_{v',i}}{{\displaystyle q^{\alpha }_{v,k}( v',i)}}\right) +{\displaystyle X_{v,k}}\sum _{v'\in \mathbf{V} \backslash v}{\displaystyle q^{\alpha }_{v,k}( v',k)}\log\left(\frac{\alpha _{v',v} \kappa ( 0) X_{v',k}}{{\displaystyle q^{\alpha }_{v,k}( v',k)}}\right)\right].
\end{aligned}
\end{equation}
Then, by setting $\frac{\partial Q(\Theta |\Theta ^{(j)} )}{\partial \mu _{v}} =0$ and $\frac{\partial Q(\Theta |\Theta ^{(j)} )}{\partial \alpha _{v',v}} =0$, we obtain the close-form iteration formulas:
\begin{equation}
\begin{aligned}
\mu ^{(j+1)}_{v} & =\frac{\sum\nolimits ^{K}_{k=1} X_{v,k} q^{\mu }_{v,k}}{K\Delta }\\
\alpha ^{(j+1)}_{v',v} & =\begin{cases}
\frac{\sum\nolimits ^{K}_{k=1}\sum ^{k}_{i=1} q^{\alpha }_{v,k} (v',i)X_{v,k}}{\sum\nolimits ^{K}_{k=1}\sum ^{k}_{i=1} \kappa ((k-i)\Delta )X_{v',i} \Delta } & v'\neq v\\
\frac{\sum\nolimits ^{K}_{k=1}\sum\nolimits ^{k-1}_{i=1} q^{\alpha }_{v,k} (v',i)X_{v,k}}{\sum\nolimits ^{K}_{k=1}\sum\nolimits ^{k-1}_{i=1} \kappa ((k-i)\Delta )X_{v',i} \Delta } & v'=v
\end{cases}
\end{aligned}
\end{equation}

\section{Proof of Theorem 2}\label{asec:proof2}
Here, to present a better understanding of the identifiability of the instantaneous causal structure, we provide two different proofs for Theorem 2 from the perspective of the likelihood and the distribution of $Y$, respectively.
\begin{customthm}{2}\label{athm:identifiability}
Let $X\to Y$ be the correct causal direction that follows
\begin{equation}\label{aeq:identifiability_model}
	\begin{aligned}[c]
	Y&=\sum\nolimits_{i=1}^{X}\xi_i + \epsilon,&\quad X, \xi_i, \textrm{ and }\epsilon \textrm{~are independent},\\
	\end{aligned}
\end{equation}
where $\xi_i\sim \operatorname{Pois}(\alpha_{X,Y})$, $\epsilon\sim \operatorname{Pois}(\mu_Y)$, $X\sim \operatorname{Pois}(\mu_X)$.
Then, there does not exist a backward model that admits the following equation:
\begin{equation}\label{aeq:identifiability_backward_model}
	\begin{aligned}[c]
	 X&=\sum\nolimits_{i=1}^{Y}\hat{\xi}_i + \hat{\epsilon}, &\quad Y,\hat{\xi}_i, \textrm{ and }\hat{\epsilon } \textrm{~are independent},
	\end{aligned}
\end{equation}
where $\hat{\xi}_i\sim \operatorname{Pois}(\hat{\alpha}_{Y,X})$, $\hat{\epsilon}\sim \operatorname{Pois}(\hat{\mu}_X)$, $Y\sim \operatorname{Pois}(\hat{\mu}_Y)$.
\end{customthm}

\subsection{Proof by Likelihood Function}
\begin{proof}
Let $\pi(x,y):=\log p(x,y)$ denote the log-likelihood of the causal model. We will prove by contradiction that there does not exist a backward model that has the same log-likelihood as the causal direction, i.e., distribution equivalent.

Suppose that there exists a backward model that has the same log-likelihood as the causal direction. Then based on the model given in Eq. \ref{aeq:identifiability_model}, the log-likelihood of the reversed direction can be written as follows. For the causal direction,
\begin{equation}\label{eq:causal_likelihood}
    \begin{aligned}
\pi(x,y) & =\log p( x) +\log p( y|x)\\
 & =x\log \mu_X -\mu_X -\log x!+y\log( x\alpha_{X,Y}+\mu_Y) -x\alpha_{X,Y}-\mu_Y -\log y!,
\end{aligned}
\end{equation}
and for the reverse direction,
\begin{equation}\label{eq:reverse_likelihood}
\begin{aligned}
\pi(x,y) & =\log p( y) +\log p( x|y)\\
 & =y\log \hat{\mu}_Y -\hat{\mu}_Y -\log y!+x\log(y\hat{\alpha}_{Y,X} +\hat{\mu}_X) -y\hat{\alpha}_{Y,X} -\hat{\mu}_X -\log x!.
\end{aligned}
\end{equation}

Let $\Delta_x \pi(x,y):= \pi(x+1,y)-\pi(x,y)$, $\Delta_{x}^2 \pi(x,y):= \Delta_{x} \pi(x+1,y)-\Delta_{x} \pi(x,y)$ denotes the first and the second order of the difference of $X$, respectively.

If there exists a backward model, then Eq. \ref{eq:reverse_likelihood} holds, implying:
\begin{equation}
   \Delta_x \pi(x,y)= \log(y\hat{\alpha}_{Y,X} +\hat{\mu}_X) -\log( x+1)
\end{equation}
and the second order of difference:
\begin{equation}\label{eq:second_diff_reverse}
   \Delta_x^2 \pi(x,y)= -\log( x+2)+\log( x+1)
\end{equation}
Using Eq. \ref{eq:causal_likelihood}, we have 
\begin{equation}
     \Delta_x \pi(x,y)= \log \mu_X -\log( x+1) +y\log( \alpha_{X,Y} ( x+1) +\mu_Y) -y\log( \alpha_{X,Y} ( x+1) +\mu_Y) -\alpha_{X,Y},
\end{equation}
and
\begin{equation}\label{eq:second_diff_causal}
     \Delta ^{2}_{x} \pi ( x,y) =-\log( x+2) +\log( x+1) +y\log( \alpha_{X,Y} ( x+2) +\mu_Y) -2y\log( \alpha_{X,Y} ( x+1) +\mu_Y) +y\log( x\alpha_{X,Y}+\mu_Y).
\end{equation}
Combining Eq. \ref{eq:second_diff_reverse} and Eq. \ref{eq:second_diff_causal}, yields
\begin{equation}\label{eq:identifi_condition}
    y\log( \alpha_{X,Y} ( x+2) +\mu_Y) -2y\log( \alpha_{X,Y} ( x+1) +\mu_Y) +y\log( x\alpha_{X,Y}+\mu_Y)=0
\end{equation}
for all $x,y$ holds. The necessary condition for Eq. \ref{eq:identifi_condition} holds for all $x,y\geq 0$ is that $\alpha_{X,Y}=0$ which contradicts the model assumption that $\alpha_{X,Y}\ne0$. This completes the proof.
\end{proof}

\subsection{Proof by Probability Generating Function}
\begin{proof}
If there exists a backward causal model following Eq. \ref{aeq:identifiability_backward_model}, then $Y$ must be the Poisson distribution, otherwise, the distribution will not be equivalent, and a simple distribution test would identify the causal direction. Thus, to show the identifiability of such a model, we only need to prove that the distribution of $Y$ can not be Poisson. 

Let $\Psi_X(s)=E\left(s^X\right)$ be the probability generating function (PGF) of the discrete random variable $X$, where $s$ belongs to some interval containing 1. Specifically, for $X\sim \operatorname{Pois}(\mu_X)$, the PGF of $X$ has the form $\Psi_X(s)= \exp[\mu_X (s-1)]$, and similarly $\Psi_{\xi_i}(s)=\exp[\alpha_{X,Y} (s-1)]$, $\Psi_{\epsilon}(s)=\exp[\mu_Y (s-1)]$. Then, based on the causal model in Eq. \ref{aeq:identifiability_model}, the probability generating function of $Y$ can be written as follows:
\begin{equation}\label{eq:distribution_y}
    \Psi_Y(s)=\Psi_X(\Psi_{\xi_i}(s))\Psi_{\epsilon}(s),
\end{equation}
which yields
\begin{equation}\label{eq:psiy1}
\Psi _{Y}( s) =\exp[ \mu _{X}(\exp[\alpha _{X,Y}( s-1)] -1)]\exp[ \mu_Y( s-1)],
\end{equation}

Suppose the backward model holds, and the desired Poisson PGF of $Y$ is that
\begin{equation}\label{eq:psiy2}
\Psi _{Y}( s) =\exp[ \hat{\mu}_{Y}( s-1)],
\end{equation}
and the necessary condition for Eq. \ref{eq:psiy1} and Eq. \ref{eq:psiy2} has the same form for all $s$ is that $\mu_X=0$ or $\alpha_{X,Y}=0$. If $\mu_X=0$, the causal variable $X$ is a constant that contradicts the model assumption. Similarly, $\alpha_{X,Y}=0$ also contradicts the assumption that $\alpha_{X,Y}\ne0$. Thus, Eq. \ref{eq:psiy1} and Eq. \ref{eq:psiy2} do not have the same form and $Y$ can not be Poisson, which completes the proof.
\end{proof}

\section{Proof of Theorem 3}\label{asec:proof3}

\begin{customthm}{3}
With the causal faithfulness assumption and causal sufficiency assumption, the multivariate instantaneous causal structure is identifiable.
\end{customthm}
\begin{proof}
With the causal faithfulness and the causal sufficiency assumption, we can identify the causal structure through conditional independence or the sparsity up to the Markov equivalent class since all Markov equivalent classes share the same skeleton. We therefore only need to show that any graph that has the same skeleton will not admit another causal structure that has a different causal direction while having the same likelihood (i.e., distribution equivalent). To show this, for a multivariate instantaneous causal structure, the log-likelihood of the causal graph $\displaystyle \mathcal{G}$ with parameters $\displaystyle \Theta $ is given as follows:
\begin{equation}
L(\mathcal{G} ,\Theta ;\mathbf{X}) =\sum _{v\in \mathbf{V}}\log p( X_{v} |X_{\mathbf{Pa}^{\mathcal{G}}_{v}}) .
\end{equation}
\begin{figure}[h]
    \centering
    \subfigure[Correct causal pair in graph $\mathcal{G}$]{
    \label{fig:theorm3a}
        \includegraphics[width=0.2\textwidth]{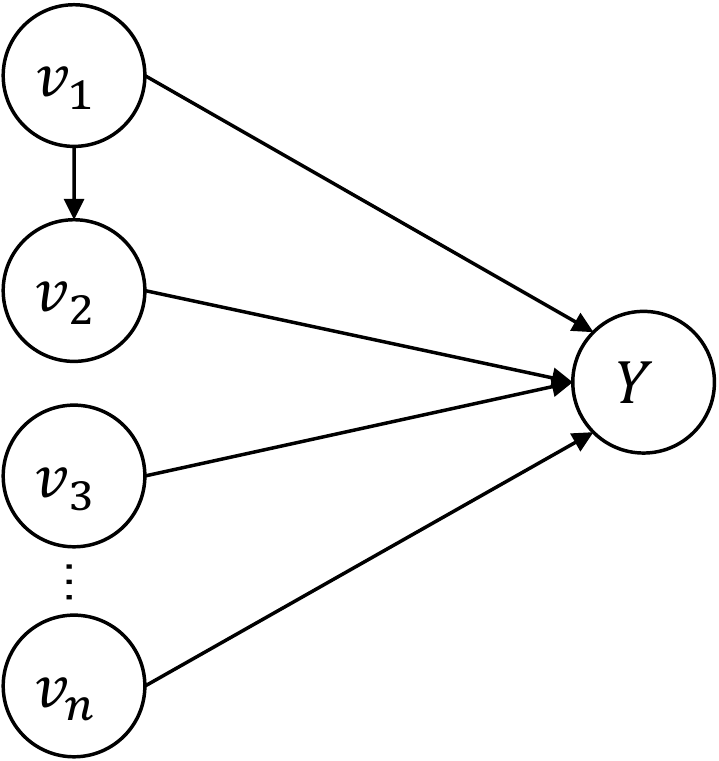}
    }
    \hspace{1cm}
    \subfigure[Reversed causal pair in graph $\hat{\mathcal{G}}$]{
    \label{fig:theorm3b}
        \includegraphics[width=0.2\textwidth]{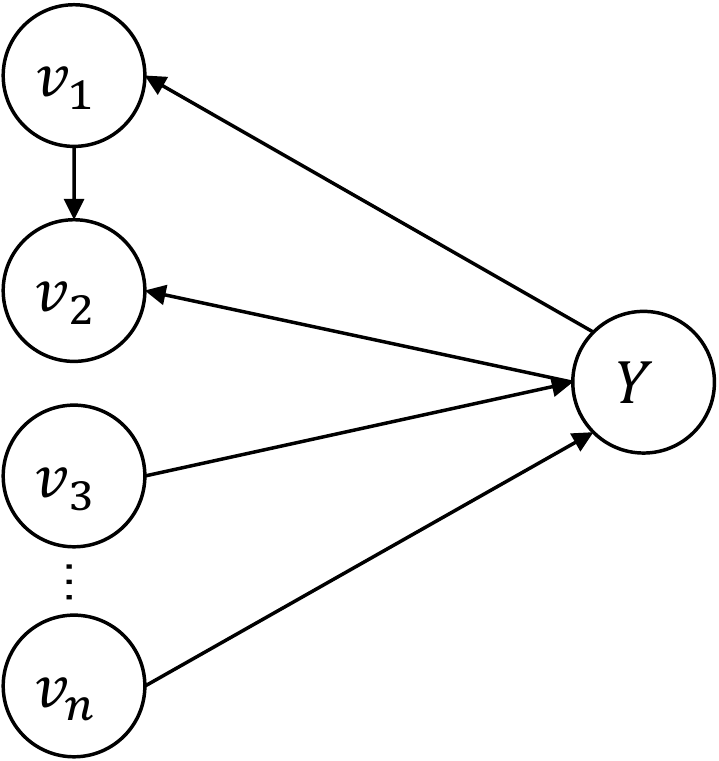}
    }
    \caption{An example for two causal graphs that has only one reversed causal pair. We let $S=\{v_1,v_2\}$ denote the set of reversed nodes in $\hat{\mathcal{G}}$ compared with $\mathcal{G}$}
    \label{fig:theorm3}
\end{figure}

Because the likelihood can be decomposed according to the graph, we can instead analyze the identifiability of each causal pair as shown in Fig. \ref{fig:theorm3}, that for the correct causal pair in Fig. \ref{fig:theorm3a} will not admit the reversed causal direction in Fig. \ref{fig:theorm3b}. Thus, the full likelihood will also not admits another causal graph among the Markov equivalent class. 

The proof is similar to Theorem \ref{athm:identifiability}, and we prove it by contradiction. Without loss of generality, as shown in Fig. \ref{fig:theorm3}, we only focus on the graph that only one causal pair has reversed edges denoted by $\displaystyle \hat{\mathcal{G}}$, while we denote the correct causal graph as $\displaystyle \mathcal{G}$. The main difference between these two graphs is the reversed nodes $\displaystyle S=\left\{v_{i} |Y\rightarrow v_{i} \in \hat{\mathcal{G}}\land Y\rightarrow v_{i}\not{\in }\mathcal{G}\right\}$ such that $\displaystyle \mathbf{Pa}^{\mathcal{G}}_{Y} =\mathbf{Pa}^{\hat{\mathcal{G}}}_{Y} \cup S$ and $\displaystyle \mathbf{Ch}^{\mathcal{G}}_{S} =\mathbf{Ch}^{\hat{\mathcal{G}}}_{S} \cup Y$. For example, in Fig. \ref{fig:theorm3}, $\displaystyle S=\{X_{1} ,X_{2}\}$. 

Suppose that the instantaneous causal structure holds with graph $\displaystyle \hat{\mathcal{G}}$ such that $\displaystyle L(\hat{\mathcal{G}} ,\hat{\Theta } ;\mathbf{X}) =L(\mathcal{G} ,\Theta ;\mathbf{X})$ then the log-likelihood of $\displaystyle \hat{\mathcal{G}}$ can be written as follows:
\begin{equation}
\pi ( X_{v_{1}} ,X_{v_{2}} ,...,X_{v_{n}}) :=\log p( X_{v_{1}} ,X_{v_{2}} ,...,X_{v_{n}}) =\sum _{v\in \mathbf{V}}\log p( X_{v} |X_{\mathbf{Pa}^{\hat{\mathcal{G}}}_{v}}) .
\end{equation}
For some set $\displaystyle S=\{v_{i} ,v_{j} ,...\}$, we denote the first-order difference of $\displaystyle \pi $ as follows:
\begin{equation}
\Delta _{S} \pi ( X_{v_{1}} ,X_{v_{2}} ,...,X_{v_{n}}) = \pi ( X_{v_{1}} ,...,X_{v_{i}} +1,...,X_{v_j} +1,...,X_{v_{n}}) -\pi ( X_{v_{1}} ,...,X_{v_{i}} ,...,X_{v_j} ,...,X_{v_{n}}) ,
\end{equation}
and the second-order difference of $\displaystyle \pi $:
\begin{equation}
\Delta ^{2}_{S} \pi ( X_{v_{1}} ,X_{v_{2}} ,...,X_{v_{n}}) =\Delta _{S} \pi ( X_{v_{1}} ,...,X_{v_{i}} +1,...,X_{v_j} +1,...,X_{v_{n}}) -\Delta _{S} \pi ( X_{v_{1}} ,...,X_{v_{i}} ,...,X_{v_j} ,...,X_{v_{n}}) .
\end{equation}
Then, by taking the first-order difference on $\displaystyle S$, we have
\begin{equation}
\begin{aligned}
\Delta _{S} \pi ( X_{v_{1}} ,X_{v_{2}} ,\dotsc ,X_{v_{n}}) = & \sum _{S_{i} \in S}[\log P(X_{S_{i}} +1\mid X_{\mathbf{Pa}_{S_{i}}^{\hat{\mathcal{G}}} \backslash S} ,X_{\mathbf{Pa}_{S_{i}}^{\hat{\mathcal{G}}} \cap S} +1)-\log P(X_{S_{i}} \mid X_{\mathbf{Pa}_{S_{i}}^{\hat{\mathcal{G}}}} )]\\
 & +\underbrace{\sum _{v\in \mathbf{Ch}_{S}^{\hat{\mathcal{G}}} \backslash S}[\log P( X_{v} \mid X_{\mathbf{Pa}_{v}^{{\mathcal{G}}} \backslash S} ,X_{\mathbf{Pa}_{v}^{{\mathcal{G}}} \cap S} +1) -\log P( X_{v} \mid X_{\mathbf{Pa}_{v}^{{\mathcal{G}}}})]}_{:=R_{1}}\\
= & \sum _{S_{i} \in S}\Biggl\{(X_{S_{i}} +1)\log\left[ \mu _{S_{i}} +\sum _{v'\in \mathbf{Pa}_{S_{i}}^{\hat{\mathcal{G}}} \backslash S} \alpha _{v',s_{i}} X_{v'} +\sum _{v'\in \mathbf{Pa}_{S_{i}}^{\hat{\mathcal{G}}} \cap S} \alpha _{v',s_{i}} (X_{v'} +1)\right]\\
 & -X_{S_{i}}\log\left[ \mu _{S_{i}} +\sum _{v'\in \mathbf{Pa}_{S_{i}}^{\hat{\mathcal{G}}} \backslash S} \alpha _{v',s_{i}} X_{v'} +\sum _{v'\in \mathbf{Pa}_{S_{i}}^{\hat{\mathcal{G}}} \cap S} \alpha _{v',s_{i}} X_{v'}\right]\\
 & -\sum _{v'\in \mathbf{Pa}_{S_{i}}^{\hat{\mathcal{G}}} \cap S} \alpha _{v',s_{i}} -\log (X_{S_{i}} +1)\Biggr\} +R_{1}
\end{aligned}
\end{equation}
Note that in the first equality, because only one causal pair has reversed edges, the parents of variables in $\displaystyle \mathbf{Ch}^{\hat{\mathcal{G}}}_{S_{i}}\backslash S$ are the same as in graph $\displaystyle \mathcal{G}$, and therefore we write $\displaystyle \mathbf{Pa}^{\mathcal{G}}_{v}$ instead of $\displaystyle \mathbf{Pa}^{\hat{\mathcal{G}}}_{v}$. Similarly, in the second equality, since there is only one causal pair difference, we have $\displaystyle \mathbf{Pa}^{\hat{\mathcal{G}}}_{S_{i}} \cap S=\mathbf{Pa}^{\mathcal{G}}_{S_{i}} \cap S$.

Furthermore, we have
\begin{equation}
\begin{aligned}
&\Delta _{S} \pi ( X_{v_{1}} ,\dotsc ,X_{v_{i}} +1,\dotsc ,X_{v_{j}} +1,X_{v_{n}})\\
&=\sum _{S_{i} \in S}\Biggl\{(X_{S_{i}} +2)\log\left[ \mu _{S_{i}} +\sum _{v'\in \mathbf{Pa}_{S_{i}}^{\hat{\mathcal{G}}} \backslash S} \alpha _{v',s_{i}} X_{v'} +\sum _{v'\in \mathbf{Pa}_{S_{i}}^{\hat{\mathcal{G}}} \cap S} \alpha _{v',s_{i}} (X_{v'} +2)\right]\\
& -(X_{S_{i}} +1)\log\left[ \mu _{S_{i}} +\sum _{v'\in \mathbf{Pa}_{S_{i}}^{\hat{\mathcal{G}}} \backslash S} \alpha _{v',s_{i}} X_{v'} +\sum _{v'\in \mathbf{Pa}_{S_{i}}^{\hat{\mathcal{G}}} \cap S} \alpha _{v',s_{i}} (X_{v'} +1)\right]\\
& -\sum _{v'\in \mathbf{Pa}_{S_{i}}^{\hat{\mathcal{G}}} \cap S} \alpha _{v',s_{i}} -\log (X_{S_{i}} +2)\Biggr\}\\
&+\underbrace{\sum _{v\in \mathbf{Ch}_{S}^{\hat{\mathcal{G}}} \backslash S}[\log P( X_{v} \mid X_{\mathbf{Pa}_{v}^{\mathcal{G}} \backslash S} ,X_{\mathbf{Pa}_{v}^{\mathcal{G}} \cap S} +2) -\log P( X_{v} \mid X_{\mathbf{Pa}_{v}^{\mathcal{G}} \backslash S} ,X_{\mathbf{Pa}_{v}^{\mathcal{G}} \cap S} +1)]}_{:=R_{2}}
\end{aligned}
\end{equation}

then for the second-order difference, we obtain
\begin{equation}\label{eq:causal_diff1}
\begin{aligned}
\Delta _{S}^{2} \pi ( X_{v_{1}} ,X_{v_{2}} ,\dotsc ,X_{v_{n}}) & =\sum _{S_{i} \in S}\Biggl\{(X_{S_{i}} +2)\log\left[ \mu _{S_{i}} +\sum _{v'\in \mathbf{Pa}_{S_{i}}^{\hat{\mathcal{G}}} \backslash S} \alpha _{v',s_{i}} X_{t,v'} +\sum _{v'\in \mathbf{Pa}_{S_{i}}^{\hat{\mathcal{G}}} \cap S} \alpha _{v',s_{i}} (X_{v'} +2)\right]\\
 &  -2(X_{S_{i}} +1)\log\left[ \mu _{S_{i}} +\sum _{v'\in \mathbf{Pa}_{S_{i}}^{\hat{\mathcal{G}}} \backslash S} \alpha _{v',s_{i}} X_{v'} +\sum _{v'\in \mathbf{Pa}_{S_{i}}^{\hat{\mathcal{G}}} \cap S} \alpha _{v',s_{i}} (X_{v'} +1)\right]\\
 &  +X_{S_{i}}\log\left[ \mu _{S_{i}} +\sum _{v'\in \mathbf{Pa}_{S_{i}}^{\hat{\mathcal{G}}} \backslash S} \alpha _{v',s_{i}} X_{v'} +\sum _{v'\in \mathbf{Pa}_{S_{i}}^{\hat{\mathcal{G}}} \cap S} \alpha _{v',s_{i}} X_{v'}\right]\\
 &  -\log (X_{S_{i}} +2)+\log (X_{S_{i}} +1)\Biggr\} +R_{2} -R_{1}
\end{aligned}
\end{equation}

For the causal direction, 

\begin{equation}\label{eq:causal_direcition_1order}
\begin{aligned}
\Delta _{S} \pi ( X_{v_{1}} ,X_{v_{2}} ,\dotsc ,X_{v_{n}}) = & \sum _{S_{i} \in S}[\log P(X_{S_{i}} +1\mid X_{\mathbf{Pa}_{S_{i}}^{\mathcal{G}} \backslash S} ,X_{\mathbf{Pa}_{S_{i}}^{\mathcal{G}} \cap S} +1)-\log P(X_{S_{i}} \mid X_{\mathbf{Pa}_{S_{i}}^{\mathcal{G}}} )]\\
 & +\sum _{v\in \mathbf{Ch}_{S}^{\mathcal{G}} \backslash S}[\log P( X_{v} \mid X_{\mathbf{Pa}_{v}^{\mathcal{G}} \backslash S} ,X_{\mathbf{Pa}_{v}^{\mathcal{G}} \cap S} +1) -\log P( X_{v} \mid X_{Pa_{v}^{\mathcal{G}}})]\\
 & \\
= & \sum _{S_{i} \in S}[\log P(X_{S_{i}} +1\mid X_{\mathbf{Pa}_{S_{i}}^{\mathcal{G}} \backslash S} ,X_{\mathbf{Pa}_{S_{i}}^{\mathcal{G}} \cap S} +1)-\log P(X_{S_{i}} \mid X_{\mathbf{Pa}_{S_{i}}^{\mathcal{G}}} )]\\
 & +\underbrace{\sum _{v\in \mathbf{Ch}_{S}^{\hat{\mathcal{G}}} \backslash S}[\log P( X_{v} \mid X_{\mathbf{Pa}_{v}^{\mathcal{G}} \backslash S} ,X_{\mathbf{Pa}_{v}^{\mathcal{G}} \cap S} +1) -\log P( X_{v} \mid X_{Pa_{v}^{\mathcal{G}}})]}_{:=R_{1}}\\
 & +[\log P( X_{Y} \mid X_{\mathbf{Pa}_{Y}^{\mathcal{G}} \backslash S} ,X_{\mathbf{Pa}_{Y}^{\mathcal{G}} \cap S} +1) -\log P( X_{Y} \mid X_{\mathbf{Pa}_{Y}^{\mathcal{G}}})]
\end{aligned}
\end{equation}

where in the second equality, because $\displaystyle \mathbf{Ch}^{\mathcal{G}}_{S_{i}} =\mathbf{Ch}^{\hat{\mathcal{G}}}_{S_{i}} \cup Y$, we can decompose the sum into $\displaystyle \mathbf{Ch}^{\hat{\mathcal{G}}}_{S_{i}}$ and $\displaystyle Y$. 

Similarly, we have
\begin{equation}\label{eq:causal_direcition_1order_plus_1}
\begin{aligned}
 & \Delta _{S} \pi (X_{v_{1}} ,\dotsc ,X_{v_{i}} +1,\dotsc ,X_{v_{j}} +1,\dotsc ,X_{v_{n}} )\\
= & \sum _{S_{i} \in S}[\log p(X_{S_{i}} +2\mid X_{\mathbf{Pa}_{S_{i}}^{\mathcal{G}} \backslash S} ,X_{\mathbf{Pa}_{S_{i}}^{\mathcal{G}} \cap S} +2)-\log p(X_{S_{i}} +1\mid X_{\mathbf{Pa}_{S_{i}}^{\mathcal{G}} \backslash S} ,X_{\mathbf{Pa}_{S_{i}}^{\mathcal{G}} \cap S} +1)]\\
 & +\underbrace{\sum _{v\in \mathbf{Ch}_{S}^{\hat{\mathcal{G}}} \backslash S}[\log P( X_{v} \mid X_{\mathbf{Pa}_{v}^{\mathcal{G}} \backslash S} ,X_{\mathbf{Pa}_{v}^{\mathcal{G}} \cap S} +2) -\log P( X_{v} \mid X_{\mathbf{Pa}_{v}^{\mathcal{G}} \backslash S} ,X_{\mathbf{Pa}_{v}^{\mathcal{G}} \cap S} +1)]}_{:=R_{2}}\\
 & +[\log P( X_{Y} \mid X_{\mathbf{Pa}_{Y}^{\mathcal{G}} \backslash S} ,X_{\mathbf{Pa}_{Y}^{\mathcal{G}} \cap S} +2) -\log P( X_{Y} \mid X_{\mathbf{Pa}_{Y}^{\mathcal{G}} \backslash S} ,X_{\mathbf{Pa}_{Y}^{\mathcal{G}} \cap S} +1)]
\end{aligned}
\end{equation}

By expanding Eq. \ref{eq:causal_direcition_1order} and Eq.  \ref{eq:causal_direcition_1order_plus_1}, we have
\begin{equation}
\begin{aligned}
\Delta _{S} \pi (X_{v_{1}} ,\dotsc ,X_{v_{n}} ) & =\sum _{S_{i} \in S}\Biggl\{(X_{S_{i}} +1)\log\left[ \mu _{S_{i}} +\sum _{v'\in \mathbf{Pa}_{S_{i}}^{\mathcal{G}} \backslash S} \alpha _{v',s_{i}} X_{v'} +\sum _{v'\in \mathbf{Pa}_{S_{i}}^{\mathcal{G}} \cap S} \alpha _{v',s_{i}} (X_{v'} +1)\right]\\
 & -X_{t,S_{i}}\log\left[ \mu _{S_{i}} +\sum _{v'\in \mathbf{Pa}_{S_{i}}^{\mathcal{G}} \backslash S} \alpha _{v',s_{i}} X_{v'} +\sum _{v'\in \mathbf{Pa}_{S_{i}}^{\mathcal{G}} \cap S} \alpha _{v',s_{i}} X_{v'}\right]\\
 & -\sum _{v'\in \mathbf{Pa}_{S_{i}}^{\mathcal{G}} \cap S} \alpha _{v',s_{i}} -\log (X_{S_{i}} +1)\Biggr\} +R_{1}\\
 & +\Biggl\{X_{Y}\log\left[ \mu _{Y} +\sum _{v'\in \mathbf{Pa}_{Y}^{\mathcal{G}} \backslash S} \alpha _{v',Y} X_{v'} +\sum _{v'\in \mathbf{Pa}_{Y}^{\mathcal{G}} \cap S} \alpha _{v',Y} (X_{v'} +1)\right]\\
 & -X_{Y}\log\left[ \mu _{Y} +\sum _{v'\in \mathbf{Pa}_{Y}^{\mathcal{G}} \backslash S} \alpha _{v',Y} X_{v'} +\sum _{v'\in \mathbf{Pa}_{Y}^{\mathcal{G}} \cap S} \alpha _{v',Y} X_{v'} \right] -\sum _{v'\in \mathbf{Pa}_{Y}^{\mathcal{G}} \cap S} \alpha _{v',Y} -\log X_{Y}\Biggr\},
\end{aligned}
\end{equation}
and
\begin{equation}
\begin{aligned}
 & \Delta _{S} \pi (X_{v_{1}} ,\dotsc ,X_{v_{i}} +1,\dotsc ,X_{v_{j}} +1,\dotsc ,X_{v_{n}} )\\
= & \sum _{S_{i} \in S}\Biggl\{(X_{S_{i}} +2)\log\left[ \mu _{S_{i}} +\sum _{v'\in \mathbf{Pa}_{S_{i}}^{\hat{\mathcal{G}}} \backslash S} \alpha _{v',s_{i}} X_{v'} +\sum _{v'\in \mathbf{Pa}_{S_{i}}^{\hat{\mathcal{G}}} \cap S} \alpha _{v',s_{i}} (X_{v'} +2)\right]\\
 & -(X_{S_{i}} +1)\log\left[ \mu _{S_{i}} +\sum _{v'\in \mathbf{Pa}_{S_{i}}^{\hat{\mathcal{G}}} \backslash S} \alpha _{v',s_{i}} X_{v'} +\sum _{v'\in \mathbf{Pa}_{S_{i}}^{\hat{\mathcal{G}}} \cap S} \alpha _{v',s_{i}} (X_{v'} +1)\right]\\
 & -\sum _{v'\in \mathbf{Pa}_{S_{i}}^{\mathcal{G}} \cap S} \alpha _{v',s_{i}} -\log (X_{S_{i}} +2)\Biggr\} +R_{2}\\
 & +\Biggl\{X_{Y}\log\left[ \mu _{Y} +\sum _{v'\in \mathbf{Pa}_{Y}^{\mathcal{G}} \backslash S} \alpha _{v',s_{i}} X_{v'} +\sum _{v'\in \mathbf{Pa}_{Y}^{\mathcal{G}} \cap S} \alpha _{v',s_{i}} (X_{v'} +2)\right]\\
 & -X_{Y}\log\left[ \mu _{Y} +\sum _{v'\in \mathbf{Pa}_{Y}^{\mathcal{G}} \backslash S} \alpha _{v',s_{i}} X_{v'}+\sum _{v'\in \mathbf{Pa}_{Y}^{\mathcal{G}} \cap S} \alpha _{v',s_{i}}( X_{v'} +1) \right] -\sum _{v'\in \mathbf{Pa}_{Y}^{\mathcal{G}} \cap S} \alpha _{v',s_{i}} -\log X_{Y}\Biggr\},
\end{aligned}
\end{equation}
respectively.

Then for the second-order difference, we have
\begin{equation}\label{eq:causal_diff2}
\begin{aligned}
 & \Delta _{S}^{2} \pi (X_{v_{1}} ,X_{v_{2}} ,\dotsc ,X_{v_{n}} )\\
 & =\Delta _{S} \pi (X_{v_{1}} ,\dotsc ,X_{v_{i}} +1,\dotsc ,X_{v_{j}} +1,\dotsc ,X_{v_{n}} )-\Delta _{S} \pi (X_{v_{1}} ,\dotsc ,X_{v_{i}} ,\dotsc ,X_{v_{j}} ,\dotsc ,X_{v_{n}} )\\
 & =\sum _{S_{i} \in S}\Biggl\{(X_{S_{i}} +2)\log\left[ \mu _{S_{i}} +\sum _{v'\in \mathbf{Pa}_{S_{i}}^{\mathcal{G}} \backslash S} \alpha _{v',s_{i}} X_{v'} +\sum _{v'\in \mathbf{Pa}_{S_{i}}^{\mathcal{G}} \cap S} \alpha _{v',s_{i}} (X_{v'} +2)\right]\\
 & -2(X_{S_{i}} +1)\log\left[ \mu _{S_{i}} +\sum _{v'\in \mathbf{Pa}_{S_{i}}^{\mathcal{G}} \backslash S} \alpha _{v',s_{i}} X_{v'} +\sum _{v'\in \mathbf{Pa}_{S_{i}}^{\mathcal{G}} \cap S} \alpha _{v',s_{i}} (X_{v'} +1)\right]\\
 & +X_{S_{i}}\log\left[ \mu _{S_{i}} +\sum _{v'\in \mathbf{Pa}_{S_{i}}^{\mathcal{G}} \backslash S} \alpha _{v',s_{i}} X_{v'} +\sum _{v'\in \mathbf{Pa}_{S_{i}}^{\mathcal{G}} \cap S} \alpha _{v',s_{i}} X_{v'}\right]\\
 & -\log (X_{S_{i}} +2)+\log (X_{S_{i}} +1)\Biggr\} +R_{2} -R_{1}\\
 & +\Biggl\{X_{Y}\log\left[ \mu _{Y} +\sum _{v'\in \mathbf{Pa}_{Y}^{\mathcal{G}} \backslash S} \alpha _{v',Y} X_{v'} +\sum _{v'\in \mathbf{Pa}_{Y}^{\mathcal{G}} \cap S} \alpha _{v',Y} (X_{v'} +2)\right]\\
 & -2X_{Y}\log\left[ \mu _{Y} +\sum _{v'\in \mathbf{Pa}_{Y}^{\mathcal{G}} \backslash S} \alpha _{v',Y} X_{v'} +\sum _{v'\in \mathbf{Pa}_{Y}^{\mathcal{G}} \cap S} \alpha _{v',Y} (X_{v'} +1)\right]\\
 & +X_{Y}\log\left[ \mu _{Y} +\sum _{v'\in \mathbf{Pa}_{Y}^{\mathcal{G}} \backslash S} \alpha _{v',Y} X_{v'} +\sum _{v'\in \mathbf{Pa}_{Y}^{\mathcal{G}} \cap S} \alpha _{v',Y} X_{v'} \right]\Biggr\}
\end{aligned}
\end{equation}

If the model is not identifiable, the second-order difference should have the same value in both causal direction and the reverse direction. Thus, combining Eq. \ref{eq:causal_diff1} and Eq. \ref{eq:causal_diff2}, we obtain the following equation
\begin{equation}\label{eq:thm3_condition}
\begin{aligned}
&X_{Y}\log\left[ \mu _{Y} +\sum _{v'\in \mathbf{Pa}_{Y}^{\mathcal{G}} \backslash S} \alpha _{v',Y} X_{v'} +\sum _{v'\in \mathbf{Pa}_{Y}^{\mathcal{G}} \cap S} \alpha _{v',Y} (X_{v'} +2)\right]
\\&-2X_{Y}\log\left[ \mu _{Y} +\sum _{v'\in \mathbf{Pa}_{Y}^{\mathcal{G}} \backslash S} \alpha _{v',Y} X_{v'} +\sum _{v'\in \mathbf{Pa}_{Y}^{\mathcal{G}} \cap S} \alpha _{v',Y} (X_{v'} +1)\right]
+X_{Y}\log\left[ \mu _{Y} +\sum _{v'\in \mathbf{Pa}_{Y}^{\mathcal{G}} \backslash S} \alpha _{v',Y} X_{v'} +\sum _{v'\in \mathbf{Pa}_{Y}^{\mathcal{G}} \cap S} \alpha _{v',Y} X_{v'} \right] =0
\end{aligned}
\end{equation}
which must hold for all possible values $\displaystyle X_{v}\in \mathbb{N}$. The necessary condition for Eq. \ref{eq:thm3_condition} holds is that for all $\displaystyle v'\in S$, $\displaystyle \alpha _{v',Y} =0$ which contradicts to the assumption that $\alpha_{v',Y}\ne0$. This finishes the proof.
\end{proof}

\section{Additional Experiments}\label{asec:exp}
\begin{figure*}[h]
	\centering
	\subfigure[Sensitivity to Time Interval]{
	\includegraphics[width=0.32\textwidth]{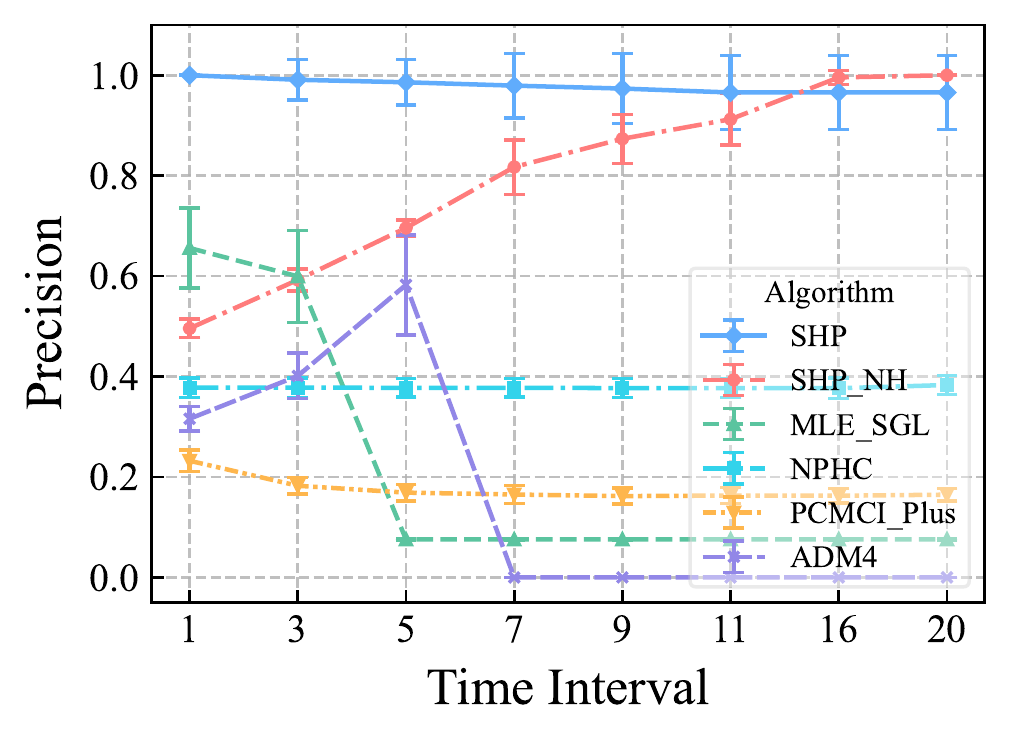}
	\label{fig:sensitivity:interval_p}
}
	\subfigure[Sensitivity to Range of $\alpha$]{
		\includegraphics[width=0.32\textwidth]{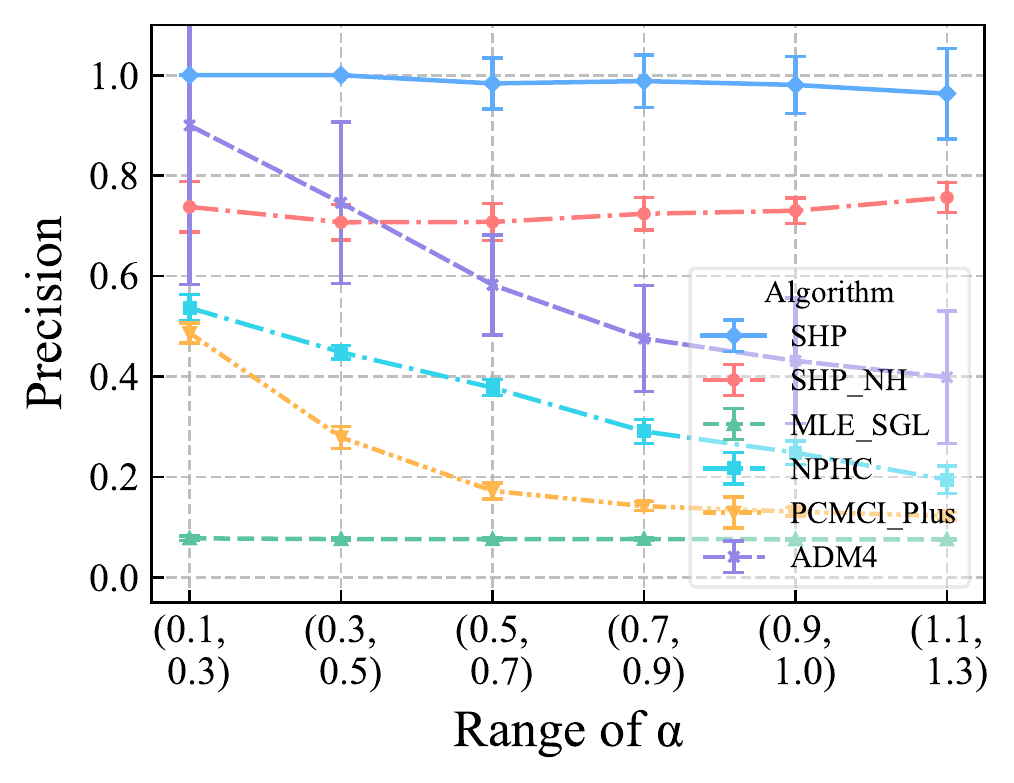}
		\label{fig:sensitivity:alpha_p}
	}
	\subfigure[Sensitivity to Range of $\mu$]{
	\includegraphics[width=0.32\textwidth]{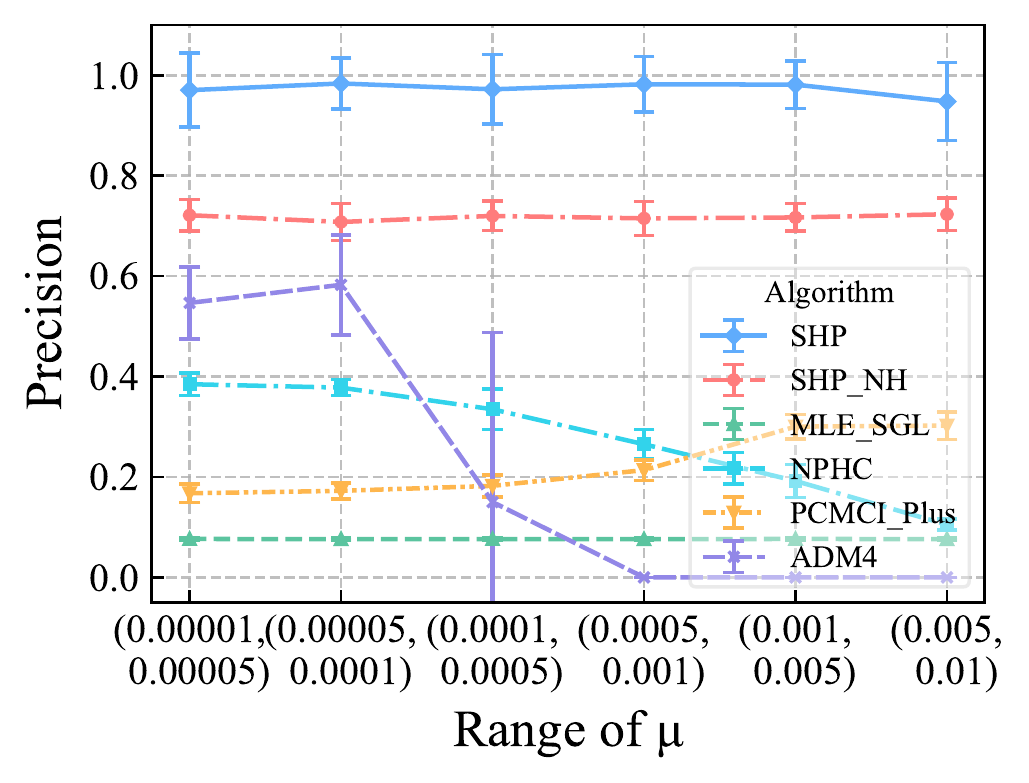}
	\label{fig:sensitivity:mu_p}
}
	\subfigure[Sensitivity to Sample Size]{
	\includegraphics[width=0.32\textwidth]{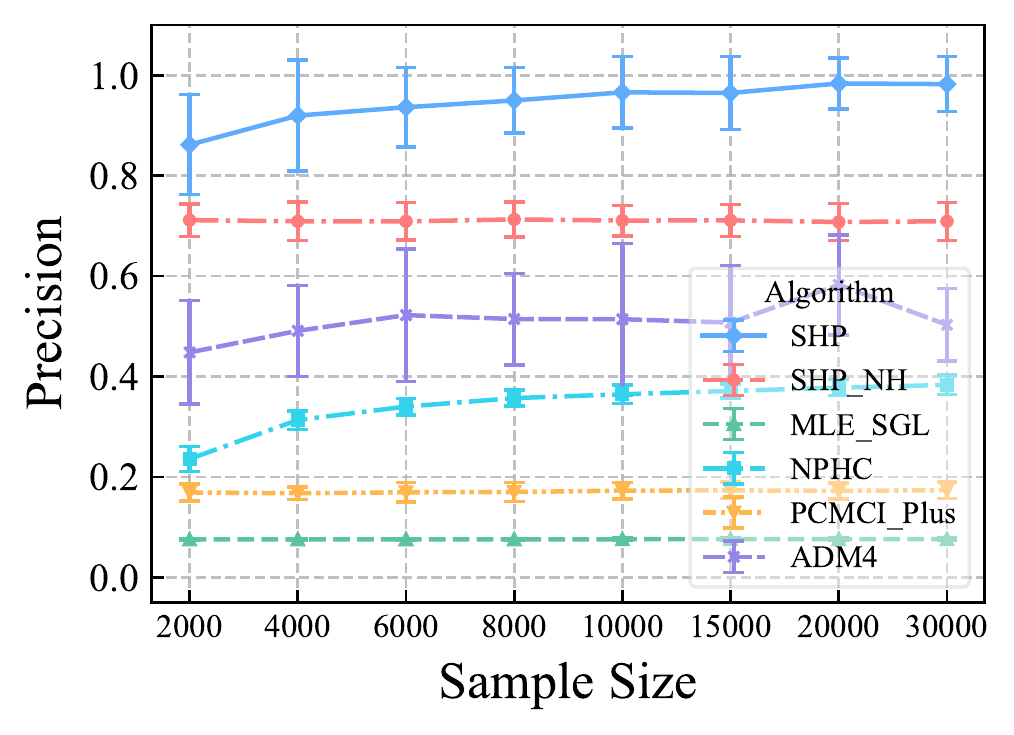}
	\label{fig:sensitivity:sample_p}
}
	\subfigure[Sensitivity to Num. of Event Types]{
	\includegraphics[width=0.32\textwidth]{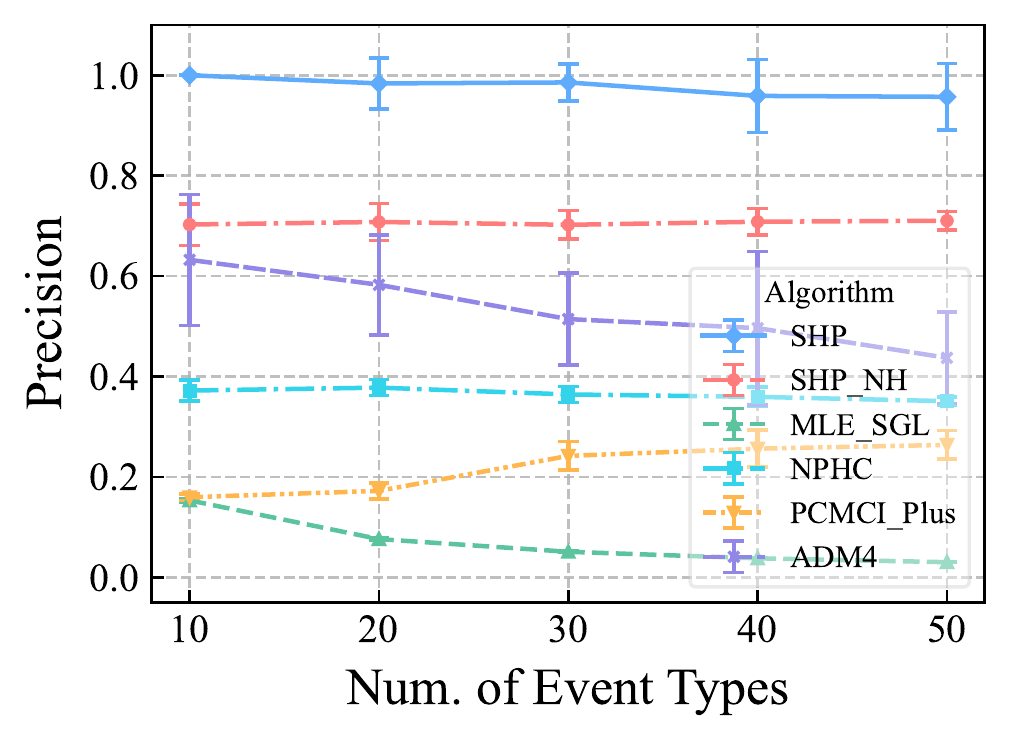}
	\label{fig:sensitivity:num_type_p}
}
	\subfigure[Sensitivity to Avg. Indegree]{
	\includegraphics[width=0.32\textwidth]{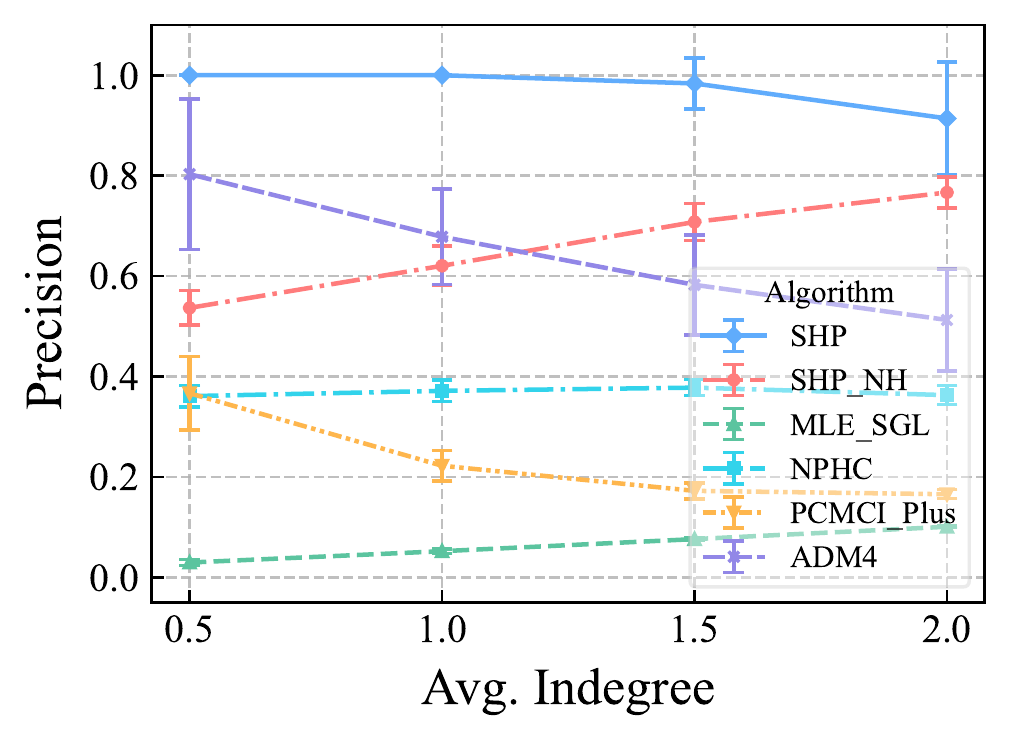}
	\label{fig:sensitivity:degree_p}
}
	\caption{Precision in the Sensitivity Experiments}	
	\label{fig:sensitivity_p}
\end{figure*}

\begin{figure*}[h]
	\centering
	\subfigure[Sensitivity to Time Interval]{
	\includegraphics[width=0.32\textwidth]{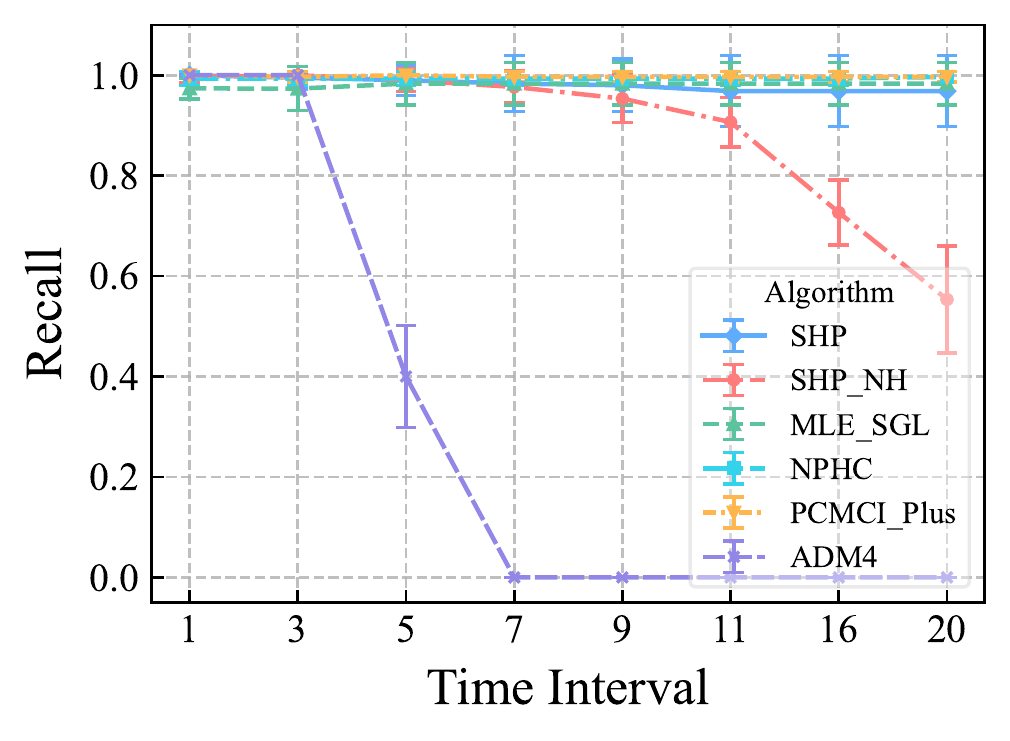}
	\label{fig:sensitivity:interval_r}
}
	\subfigure[Sensitivity to Range of $\alpha$]{
		\includegraphics[width=0.32\textwidth]{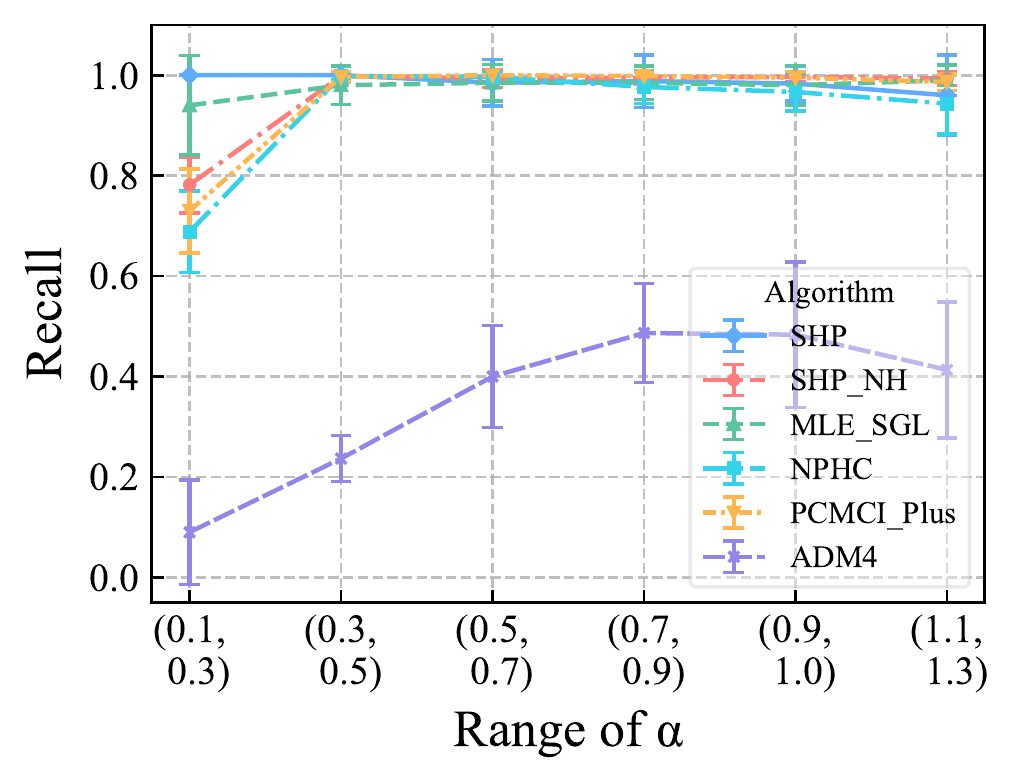}
		\label{fig:sensitivity:alpha_r}
	}
	\subfigure[Sensitivity to Range of $\mu$]{
	\includegraphics[width=0.32\textwidth]{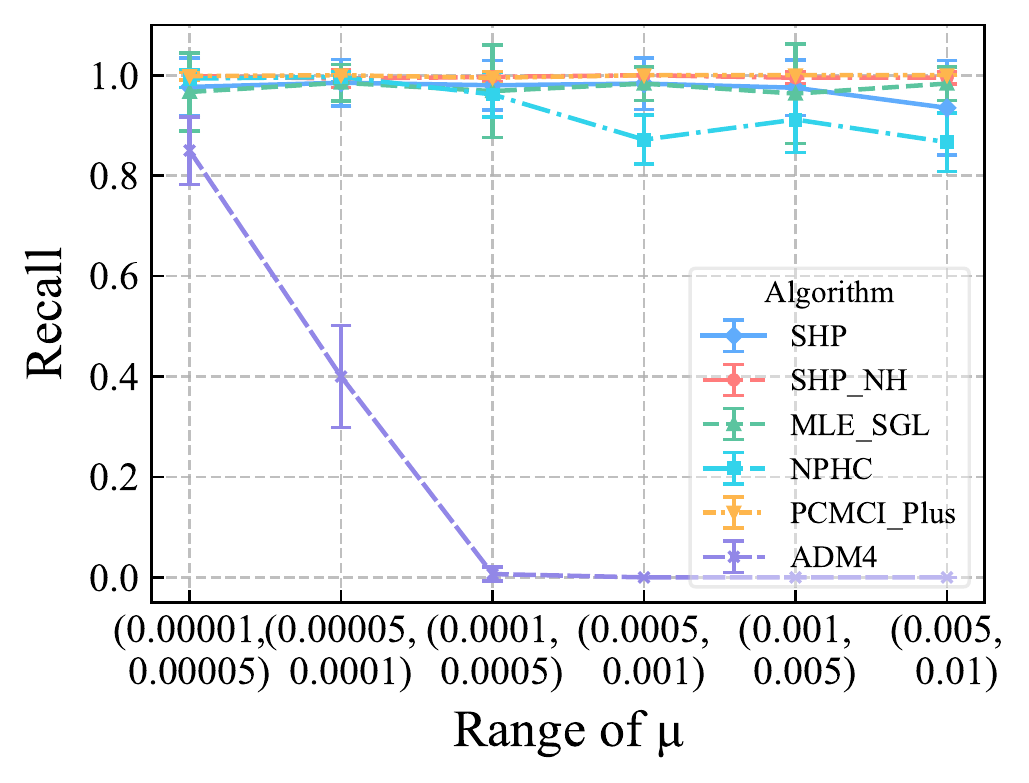}
	\label{fig:sensitivity:mu_r}
}
	\subfigure[Sensitivity to Sample Size]{
	\includegraphics[width=0.32\textwidth]{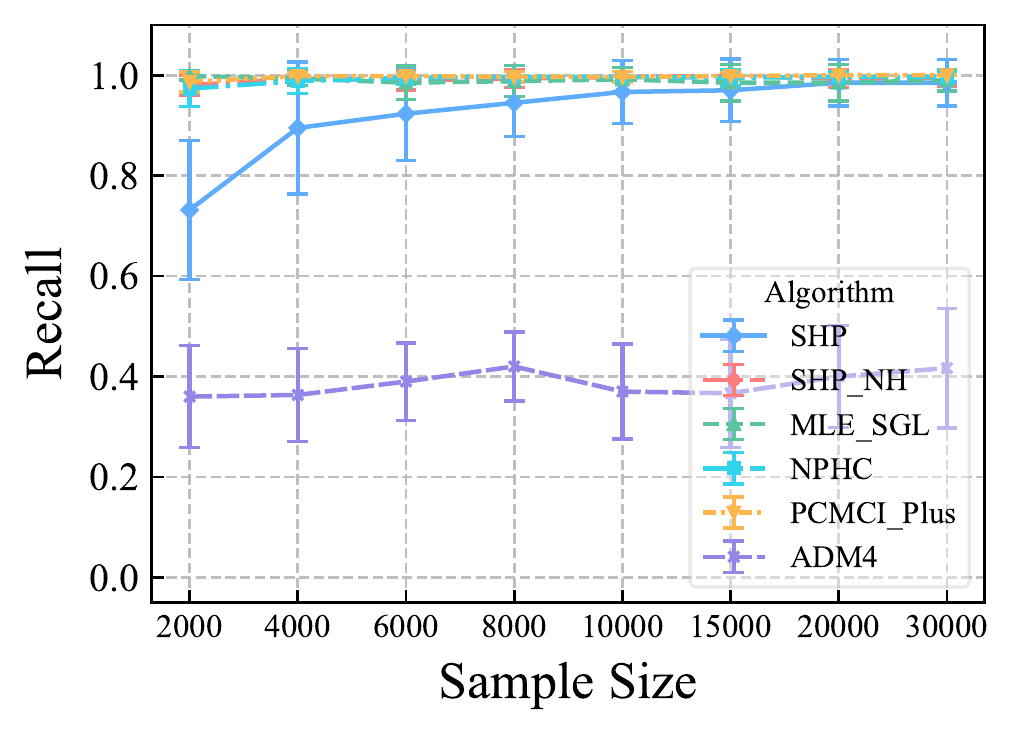}
	\label{fig:sensitivity:sample_r}
}
	\subfigure[Sensitivity to Num. of Event Types]{
	\includegraphics[width=0.32\textwidth]{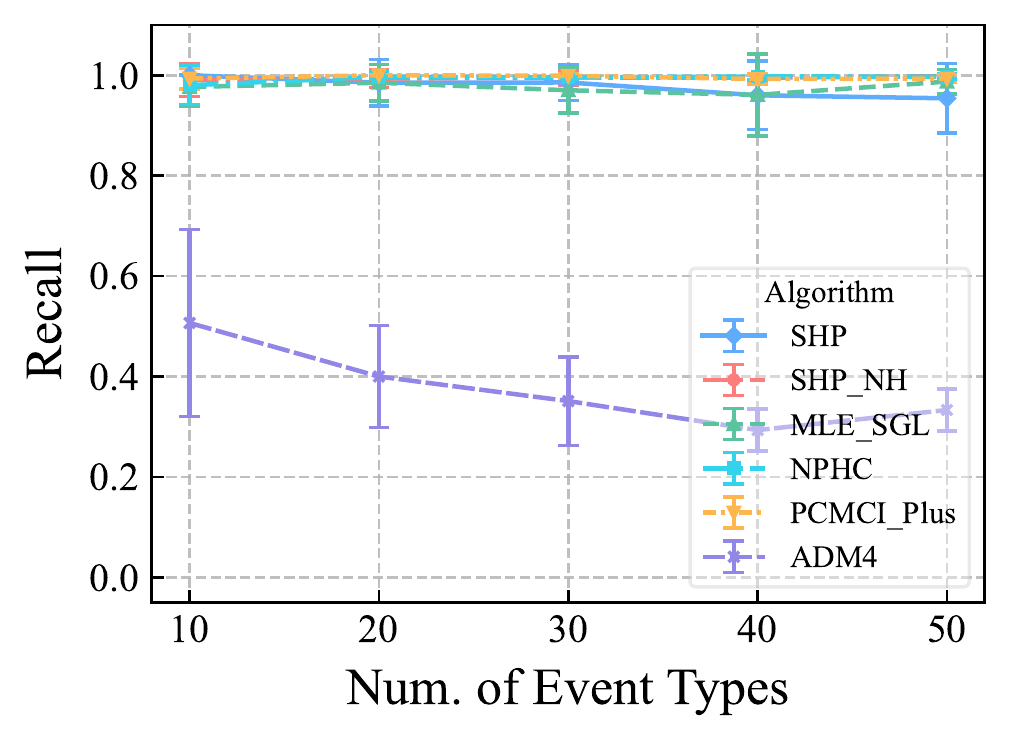}
	\label{fig:sensitivity:num_type_r}
}
	\subfigure[Sensitivity to Avg. Indegree]{
	\includegraphics[width=0.32\textwidth]{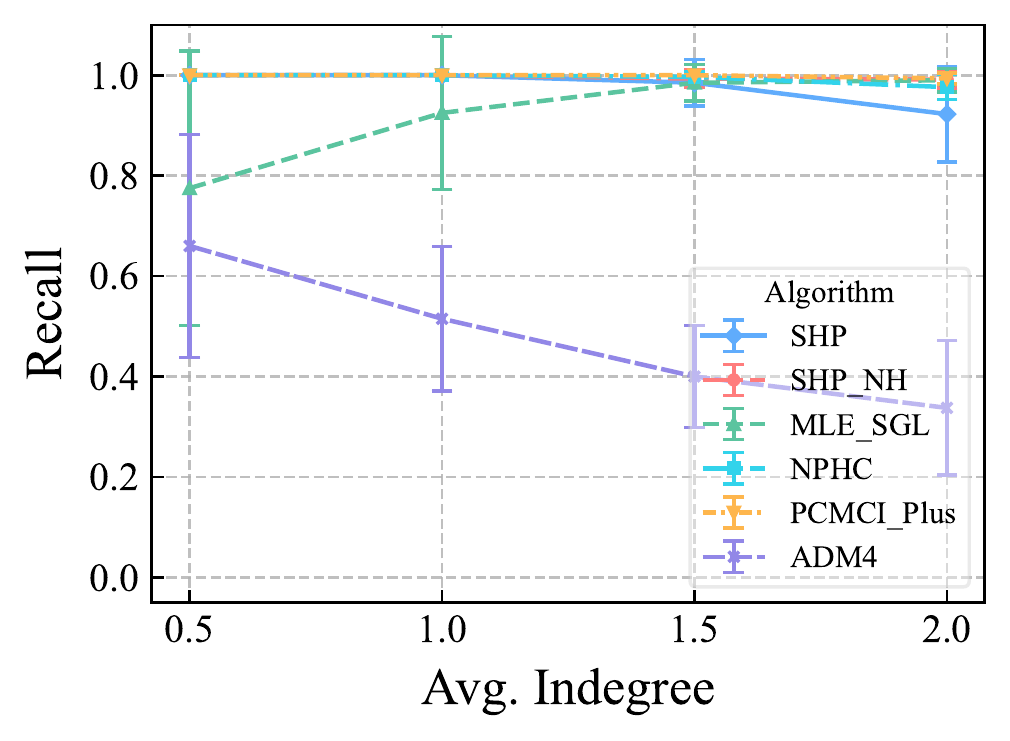}
	\label{fig:sensitivity:degree_r}
}
	\caption{Recall in the Sensitivity Experiments}	
	\label{fig:sensitivity_r}
\end{figure*}

\begin{figure*}
	\centering
	\subfigure[Sensitivity to Time Interval]{
	\includegraphics[width=0.32\textwidth]{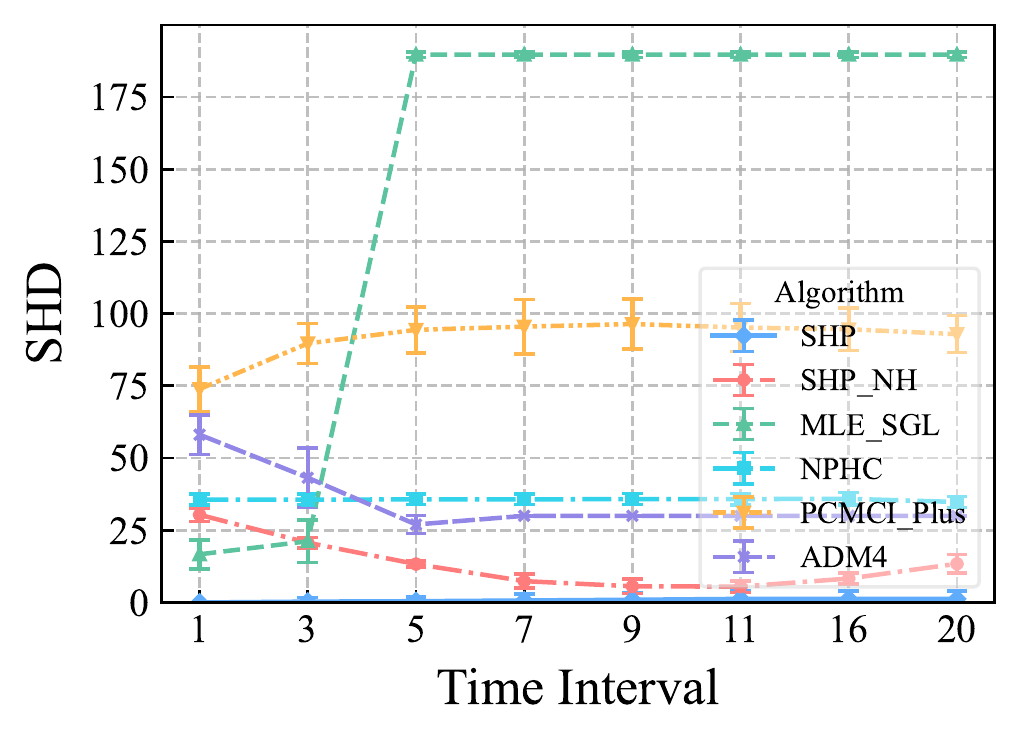}
	\label{fig:sensitivity:interval_shd}
}
	\subfigure[Sensitivity to Range of $\alpha$]{
		\includegraphics[width=0.32\textwidth]{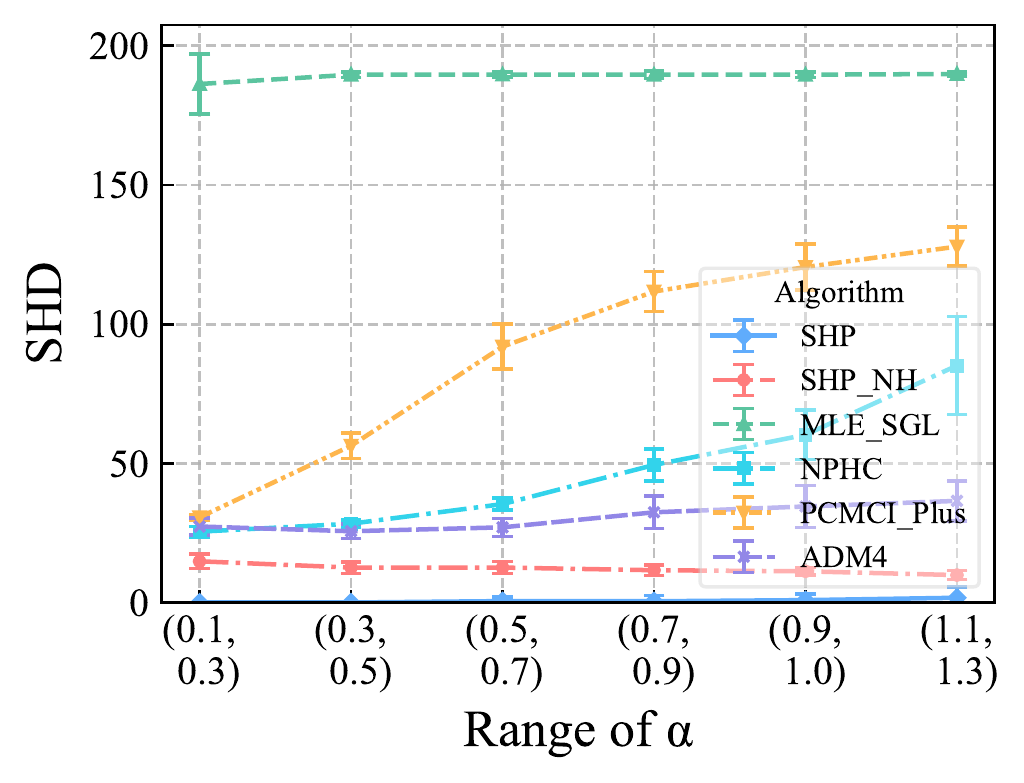}
		\label{fig:sensitivity:alpha_shd}
	}
	\subfigure[Sensitivity to Range of $\mu$]{
	\includegraphics[width=0.32\textwidth]{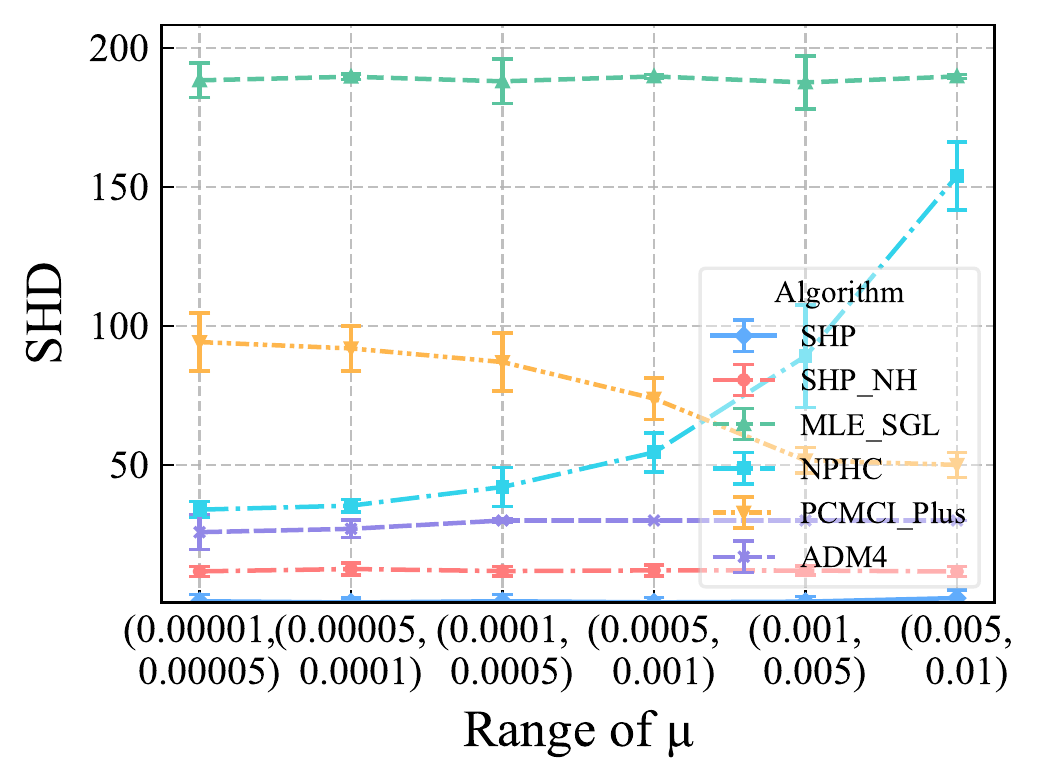}
	\label{fig:sensitivity:mu_shd}
}
	\subfigure[Sensitivity to Sample Size]{
	\includegraphics[width=0.32\textwidth]{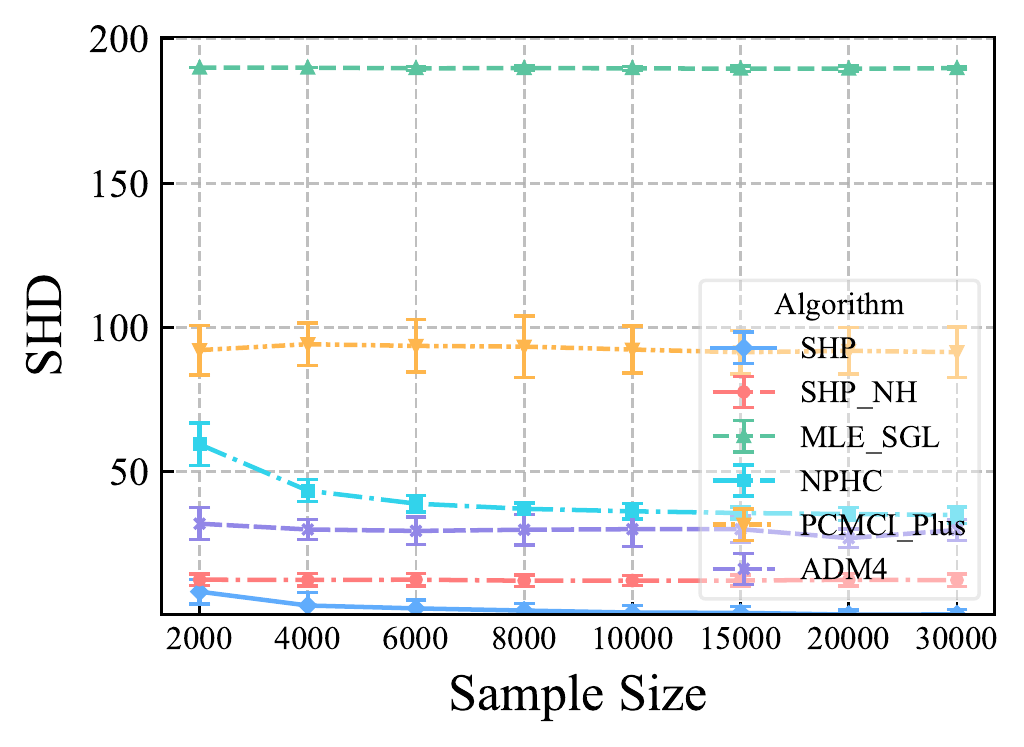}
	\label{fig:sensitivity:sample_shd}
}
	\subfigure[Sensitivity to Num. of Event Types]{
	\includegraphics[width=0.32\textwidth]{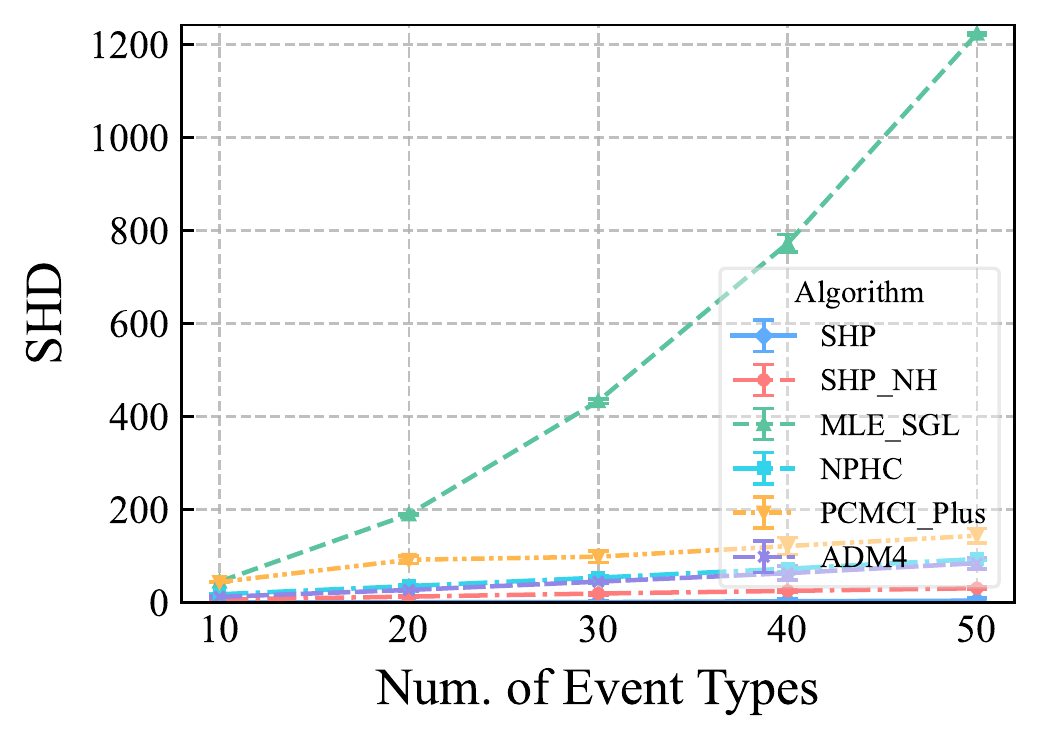}
	\label{fig:sensitivity:num_type_shd}
}
	\subfigure[Sensitivity to Avg. Indegree]{
	\includegraphics[width=0.32\textwidth]{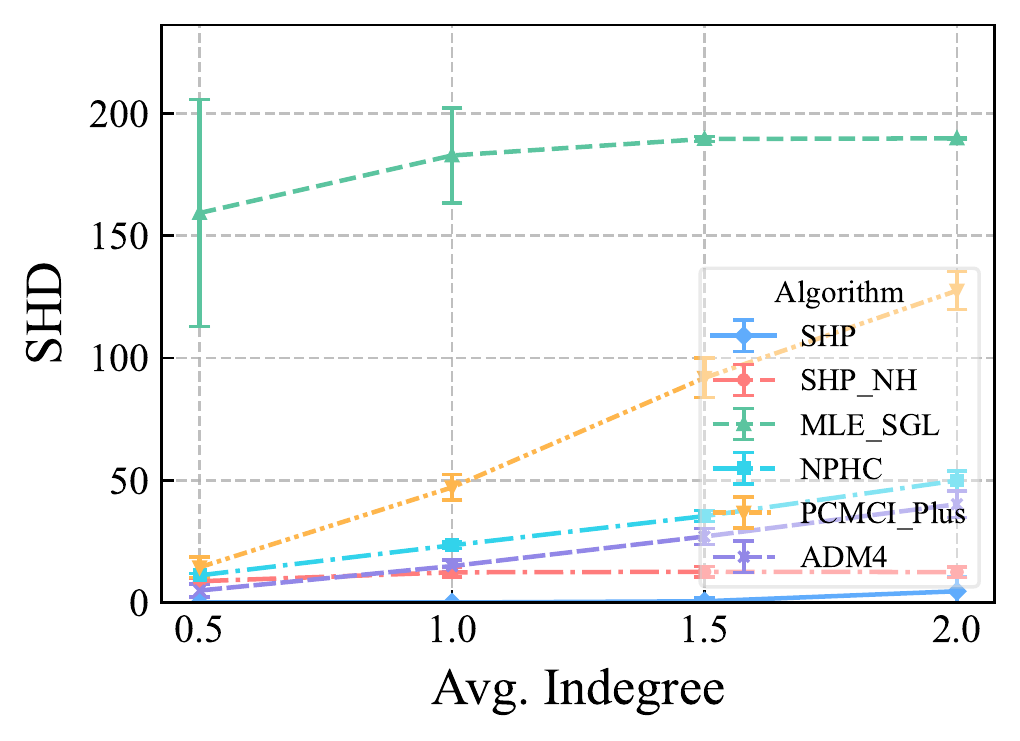}
	\label{fig:sensitivity:degree_shd}
}
	\caption{SHD in the Sensitivity Experiments}	
	\label{fig:sensitivity_shd}
\end{figure*}

\begin{figure*}[h]
	\centering
	\subfigure[Precision]{
	\includegraphics[width=0.32\textwidth]{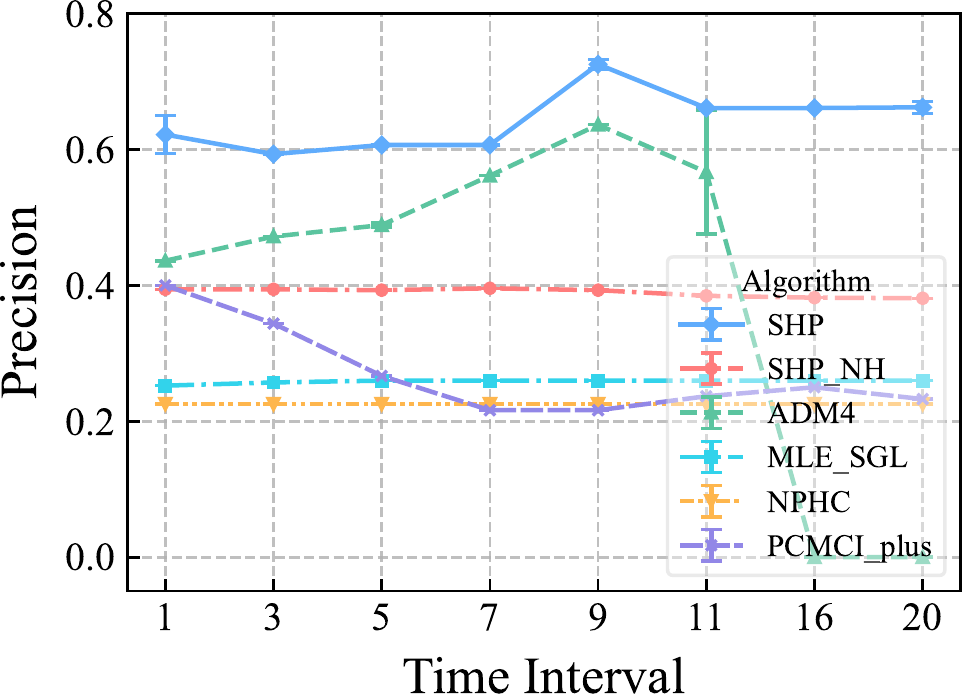}
	\label{fig:real_world_time_interval:precision}
}
	\subfigure[Recall]{
		\includegraphics[width=0.32\textwidth]{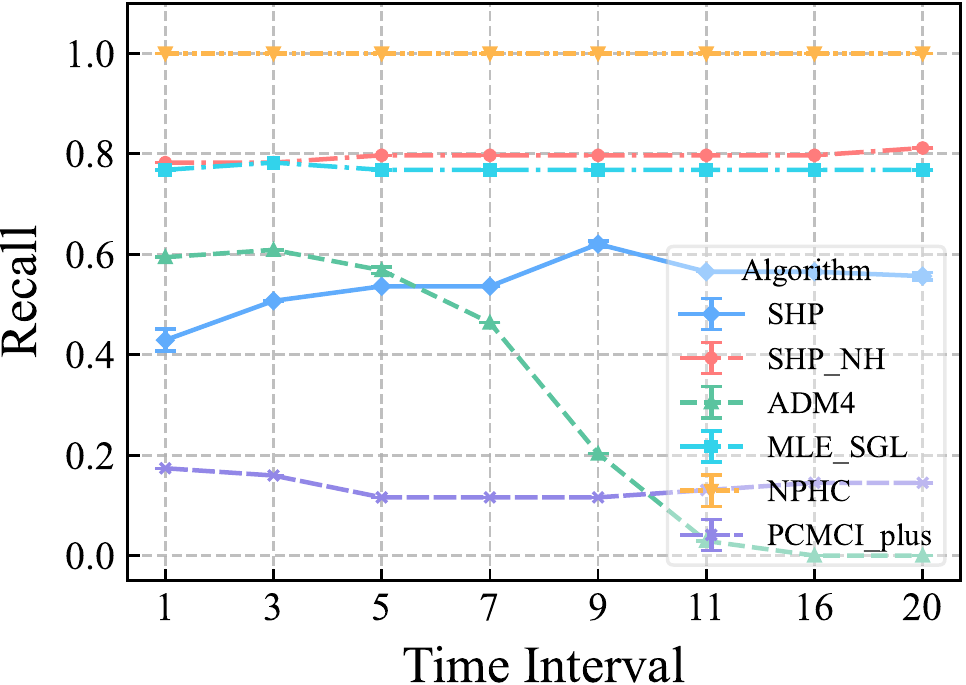}
		\label{fig:real_world_time_interval:recall}
	}
 	\subfigure[SHD]{
		\includegraphics[width=0.32\textwidth]{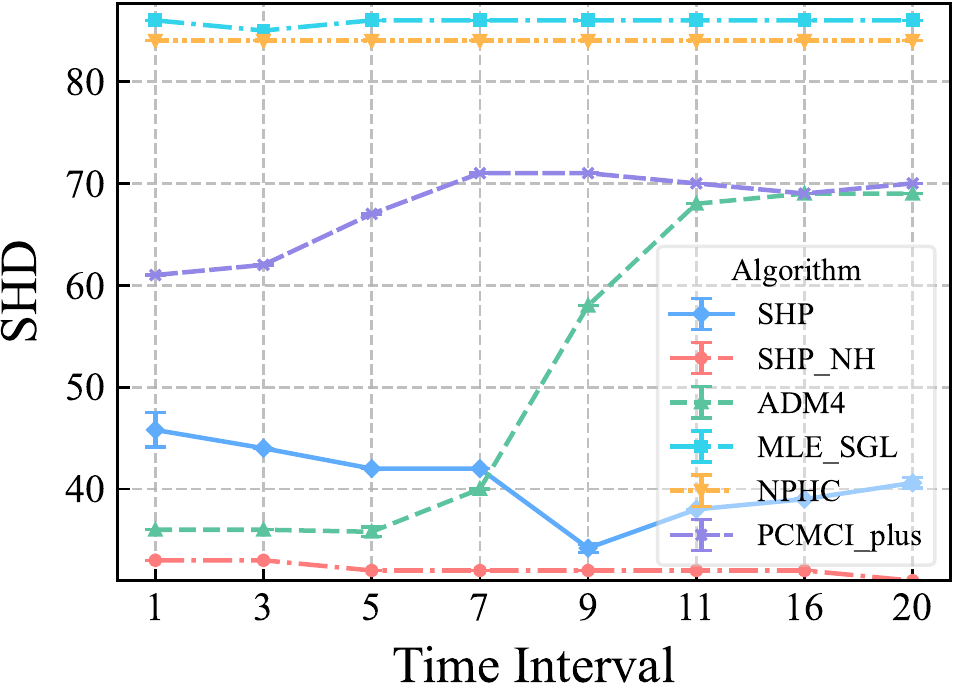}
		\label{fig:real_world_time_interval:shd}
	}
	\caption{Real-World Experiment on Different Temporal Resolutions}	
	\label{afig:real_world_time_interval}
\end{figure*}

The main paper has shown the F1 scores and other baselines in both synthetic data and real-world experiments. Here, we further provide the Precision, Recall, and Structural Hamming Distance (SHD) in these experiments, as shown in Fig. \ref{fig:sensitivity_p} and Fig. \ref{fig:sensitivity_r}, Fig. \ref{fig:sensitivity_shd}, and Fig. \ref{afig:real_world_time_interval}. 
Note that most of the parameters are based on the default setting in the tick packages \cite{JMLR:v18:17-381}.

\end{document}